\documentclass[11pt,a4paper]{article}
\usepackage{a4wide}
\usepackage{graphicx}
\usepackage[fleqn]{amsmath}
\usepackage{amssymb}
\usepackage{amsthm}
\usepackage{bm}
\usepackage{url}
\usepackage{color}

\usepackage{accents}

\theoremstyle{plain}

\theoremstyle{definition}

\theoremstyle{remark}

\allowdisplaybreaks
\graphicspath{{images/}}
\begin{document}
\title{Graph Entropies in Texture Segmentation of Images}
\author{Martin Welk\\ University for Health Sciences, Medical Informatics and Technology (UMIT),\\ Eduard-Walln\"ofer-Zentrum 1, 6060 Hall/Tyrol, Austria\\ \url{martin.welk@umit.at} }
\date{December 28, 2015}
\maketitle

\begin{abstract}
We study the applicability of a set of texture descriptors introduced in
recent work by the author to texture-based segmentation of images. The texture 
descriptors under investigation result from applying graph indices from 
quantitative graph theory to graphs encoding the local structure of images. 
The underlying graphs arise from the computation of morphological amoebas as 
structuring elements for adaptive morphology, either as weighted or unweighted
Dijkstra search trees or as edge-weighted pixel graphs within structuring
elements. In the present paper we focus on texture descriptors in which the
graph indices are entropy-based, and use them in a geodesic active contour
framework for image segmentation. Experiments on several synthetic and one
real-world image are shown to demonstrate texture segmentation by this
approach. Forthermore, we undertake an attempt to analyse selected 
entropy-based texture descriptors with regard to what information about texture
they actually encode. Whereas this analysis uses some heuristic assumptions,
it indicates that the graph-based texture descriptors are related to 
fractal dimension measures that have been proven useful in texture analysis.

\bigskip \noindent%
\end{abstract}

\section{Introduction}
\label{sec-intro}

Graph models have been used in image analysis for a long time.
The edited book
\cite{Lezoray-Book12} gives an overview of methods in this field.
However, approaches from quantitative graph theory such as
graph indices have not played a significant role in these applications so far.
This is to some extent surprising as it is not a far-fetched idea to
model information contained in small patches of a textured image by graphs,
and once this has been done, graph indices with their ability
to extract in a quantitative form structural information from large
collections of graphs lend themselves as a promising tool specifically for
texture analysis. A first step in this direction has been made in
\cite{Welk-qgt14} where a set of texture descriptors was introduced
that combines a construction of graphs from image patches with well-known
graph indices. This set of texture descriptors was evaluated in
\cite{Welk-qgt14} in the context of a texture discrimination task.
In \cite{Welk-bookchapter15x}, an example for texture-based image
segmentation was presented based on this work.

The present paper continues the work begun in
\cite{Welk-qgt14} and \cite{Welk-bookchapter15x}.
Its purpose is twofold. On one hand, it restates and slightly extends
the experimental work from \cite{Welk-bookchapter15x} on texture
segmentation, focussing on those descriptors that are based on entropy
measures, which turned out particularly useful in the previous contributions.
On the other hand, it undertakes a first attempt to analyse the graph-index
based texture descriptors with regard to what kind of information they
actually extract from a textured image.

\paragraph{Structure of the paper.}
The remaining part of Section~\ref{sec-intro} briefly outlines
the fields of research that are combined in this work,
namely quantitative graph theory, see
Section~\ref{ssec-introqgt}, graph models in image analysis with
emphasis on the pixel graph and its edge-weighted variant, see
Section~\ref{ssec-introgmia}, and finally texture analysis,
Section~\ref{ssec-introtex}.
In Section~\ref{sec-gi} the construction of graph-entropy-based
texture descriptors from \cite{Welk-qgt14} is detailed.
Section~\ref{sec-gac} gives a brief account of the geodesic active
contour method, a well-established approach for image segmentation that is
based on the numerical solution of a partial differential equation.
Texture segmentation combining the graph-entropy-based texture descriptors
with the geodesic active contour method is demonstrated on two synthetic
examples that represent typical realistic texture segmentation tasks,
and a real-world example in Section~\ref{sec-segex}.
Some theoretical analysis is presented in Section~\ref{sec-tdana}
where (one setup of) graph-entropy-based texture descriptors is put into
relation with fractal dimension measurements on a metric space derived from
the pixel graph, and thus a connection is established between graph entropy
methods and fractal-based texture descriptors.
A short conclusion, Section~\ref{sec-conc}, ends the paper.

\subsection{Quantitative Graph Theory}
\label{ssec-introqgt}

Quantitative measures for graphs have been developed for almost sixty years
in mathematical chemistry as a means to analyse molecular graphs
\cite{Bonchev-JCP77,Hosoya-BCSJ71,Ivanciuc-JMC93,
Plavsic-JMC93,Wiener-JACS47}.
In the course of time, numerous \emph{graph indices} have been derived based
on edge connectivity, vertex degrees, distances, and information-theoretic
concepts, see e.g.\ the work \cite{Dehmer-PLO13} that classifies
over $900$ descriptors from literature and subjects them to a
large-scale statistical evaluation on several test data sets.
Recently, interesting new graph indices based on the so-called
Hosoya polynomial have been proposed \cite{Dehmer-AMC15}.
Fields of application have diversified in the last decades to include
e.g.\ biological and social networks, and other structures that can
mathematically be modelled as graphs, see \cite{Dehmer-Book12}.
Efforts to apply statistical methods to graph indices
across these areas have been bundled in the emerging field of
\emph{quantitative graph theory}
\cite{Dehmer-Book14,Dehmer-Book12}.

Many contributions in this field group around the tasks of distinguishing and
classifying graphs, and quantifying the differences between graphs.
The first task focusses on the ability of indices to \emph{uniquely}
distinguish large sets of individual graphs, termed
\emph{discrimination power} \cite{Bonchev-JCC81,Dehmer-PLO12a,Dehmer-AMC15}.
For the latter task, \emph{inexact graph matching,}
the \emph{graph edit distance}
\cite{Ferrer-bookchapter12,Sanfeliu-TSMC83} or other
measures quantifying the size of substructures that are shared or
not shared between two graphs are of particular importance, see also
\cite{Cross-PR97,
Dehmer-AMC07a,Dehmer-AMC06,EmmertStreib-AMC07a,
Riesen-IVC09,Wang-IS95,Zhu-dasfaa11}.
The concept of discrimination
power has to be complemented for this purpose by the principle of
\emph{smoothness} of measures, see \cite{EmmertStreib-PLO12}, which
describes how similar the values of a measure are when it is applied to
similar graphs.
In \cite{Furtula-AMC13}, the quantitative measures of
\emph{structure sensitivity} and \emph{abruptness} have been introduced in
order to precisely analyse the discrimination power and smoothness of
graph indices. These measures are based on the average and maximum,
respectively, of the changes of graph index values when the underlying
graph is modified by one elementary edit step of the graph edit distance.

Discrimination of graph structures by graph indices is also a crucial
part of the texture analysis approach discussed here. Thus, discrimination
power and the related notions of high structure sensitivity and low
abruptness matter also in the present context.
However, unique identification of individual graphs is somewhat less
important in texture analysis than when examining single graphs as in
\cite{Dehmer-AMC15,Furtula-AMC13}, as in texture
analysis one is confronted with
large collections of graphs associated with image pixels, and one
is interested in separating these into a small number of
classes representing regions. Not only will each class contain numerous
graphs, but also the spatial arrangement of the associated pixels is to be
taken into account as an additional source of information, as segments are
expected to be connected.

\subsection{Graph Models in Image Analysis}
\label{ssec-introgmia}

As can be seen in \cite{Lezoray-Book12}, there are several ways in
which image analysis can be linked to graph concepts.
A large class of approaches is based on graphs in which the pixels of a
digital image take the role of vertices, and the set of edges is based on
neighbourhood relations, with $4$- and $8$-neighbourhoods as most popular
choices in 2D, and similar constructions in 3D, see
\cite[Section~1.5.1]{Lezoray-bookchapter12}.
To imprint actual image information on such a graph, one can furnish it
with edge weights that are based on image contrasts.
Among others, the \emph{graph cut} methods \cite{Ishikawa-bookchapter12}
that have recently received much attention for applications such as
image segmentation \cite{Ishikawa-cvpr98,Shi-PAMI00}
and correspondence problems \cite{Boykov-cvpr98,Roy-iccv98}
make use of this concept. This setup is also central for the work presented
here, see the more detailed account of the pixel graph and edge-weighted
pixel graph of an image in Section~\ref{ssec-graphcons}
of the present paper.

Generalising the pixel-graph framework, the graph perspective allows to
transfer image
processing methods from the regular mesh underlying standard digital images
to non-regular meshes that can be related to scanned surface data
\cite{Clarenz-CG04} but arise also from standard images when considering
non-local models \cite{Buades-cvpr05} that have recently received
great attention in image enhancement. Graph morphology, see e.g.\
\cite{Najman-bookchapter12}, is one of these generalisations of
image processing methods to non-regular meshes, but also variational
and PDE frameworks have been generalised in this way
\cite{Elmoataz-bookchapter12}.

We briefly mention that graphs can also be constructed, after suitable
preprocessing, from vertices representing image regions, see
\cite[Section~1.5.2]{Lezoray-bookchapter12}, opening avenues to
high-level semantic image interpretation by means of partition trees.
Comparison of hierarchies of semantically meaningful partitions can then
be achieved e.g.\ using graph edit distance or related concepts
\cite{Ferrer-bookchapter12}.

Returning to the pixel-graph setup which we will also use in this work,
see Section~\ref{ssec-graphcons}, we point out a difference of our
approach to those that represent the entire image in a single pixel graph.
We focus here on subgraphs related to small image patches, thus generating
large sets of graphs whose vertex sets, connectivity and/or edge weights
encode local image information. To extract meaningful information from such
collections, statistical methods such as entropy-based graph indices are
particularly suited.

\subsection{Texture}
\label{ssec-introtex}

In image analysis, the term \emph{texture} refers to the small-scale structure
of image regions, and as such it has been an object of intensive investigation
since the beginnings of digital image analysis.
For example,
\cite{Haralick-PIEEE79,Haralick-TSMC73,Rosenfeld-TC71,
Sutton-TC72,Zucker-CGIP76} undertook approaches to define and
analyse textures.

\paragraph{Complementarity of texture and shape.}
Real-world scenes often consist of collections of distinct
objects which in the process of imaging are mapped to regions in an image
delineated by more or less sharp boundaries. While the geometric description
of region boundaries is afforded by the concept of \emph{shape} and typically
involves large-scale structures, texture represents the appearance of the
individual objects, either their surfaces if images in the case of reflection
imaging (such as photography of opaque objects), or their interior if
transmission-based imaging modalities (such as transmission microscopy,
X-ray, magnetic resonance) are being considered. Texture is then expressed
in the distribution of intensities and their short-scale correlations
within a region. A frequently used mathematical formulation of this
distinction is the \emph{cartoon-texture model} that
underlies many works on image restoration and enhancement,
see e.g.\ \cite{Osher-MMS03}. In this approach, (space-continuous)
images are described as the sum of two functions:
a \emph{cartoon} component from the space $\mathrm{BV}$ of functions of
bounded variation, and a \emph{texture} component from a suitable Sobolev
space. In a refined version of this decomposition \cite{Aujol-IJCV05},
noise is modelled as a third component assigned to a different function
space.

Note that also in image synthesis (computer graphics) the
complementarity of shape and texture is used: Here, textures are understood
as intensity maps that are mapped on the surfaces of geometrically described
objects.

The exact frontier between shape and texture information in a scene or image,
however, is model-dependent. The intensity variation of a surface
is partly caused by geometric details of that surface. With a coarse-scale
modelling of shape, small-scale variations are included in the texture
description, whereas with a refined modelling of shape, some of these
variations become part of the shape information. For example, in the
texture samples shown in Figure~\ref{fi-e3}(a) and (b), a
geometric description with sufficiently fine granularity could capture
individual leaves or blossoms as shapes, whereas the large-scale viewpoint
treats the entire ensemble of leaves or blossoms as texture.

\paragraph{Texture models.}
Capturing texture is virtually never possible on the basis of a single
pixel. Only the simplest of all textures, homogeneous intensity, can be
described by a single intensity. For all other textures, intensities within
neighbourhoods of suitable size (that differs from texture to texture) need
to be considered to detect and classify textures. Moreover, there is a
large variety of structures that can be constitutive of textures, ranging
from periodic patterns in which the arrangement of intensity values follows
strict geometric rules, via near-periodic and quasi-periodic structures to
irregular patterns where just statistical properties of intensities within
a neighbourhood are characteristic of the texture. The texture samples in
Figure~\ref{fi-e3}(a) and (b) are located rather in the middle
of the scale where both geometric relations and statistics of the intensities
are characteristic of the texture; near-periodic stripe patterns as in the
later examples, Figures~\ref{fi-e-stripes} and \ref{fi-zebra},
are more geometrically dominated.

With emphasis on different categories of textures within this continuum,
numerous geometric and statistic approaches have been made over the decades
to describe textures. For example, frequency-based models
\cite{Gabor-JIEE46,Lendaris-PIEEE70,Sandberg-tr-cam02-39}
emphasise the periodic or quasi-periodic aspect of textures.
Statistics on intensities such as \cite{Rosenfeld-TC71} mark the
opposite end of the scale, whereas models based on statistics of
image derivative quantities like
gradients \cite{Haralick-TSMC73} or structure tensor entries
\cite{Brox-IVC10} attempt to combine statistical with geometrical
information.
A concept that differs significantly from both approaches has been proposed in
\cite{Lu-CGIP78} where textures are described generatively via grammars.

Also fractals \cite{Mandelbrot-Book77} have been proposed as a means
to describe, distinguish and classify textures. Remember that a fractal is a
geometric object, in fact a topological space, for which it is possible to
determine, at least locally, a Minkowski dimension (or, almost identical,
Hausdorff dimension) which differs from its topological dimension.
Assume that the fractal is embedded in a surrounding
Euclidean space, and it is compact. Then it can be covered by a finite number
of boxes, or balls, of prescribed size. When the size of the boxes or balls
is sent to zero, the number of them which is needed to cover the structure
grows with some power of the inverse box or ball size. The Minkowski dimension
of the fractal is essentially the exponent in this power law.
The Minkowski dimension of a fractal is often non-integer (which is the reason
for its name), however, a more precise definition is that Minkowski and
topological dimensions differ, which also includes cases like the Peano curve
whose Minkowski dimension is an integer ($2$) but still different from the
topological dimension ($1$). See also
\cite{Avadhanam-msc93,Barbaroux-JMPA01}
for different concepts of fractal dimension.

Textured images can be associated with fractals by considering the image
manifold, i.e.\ the function graph if the image is modelled as a function
over the image plane, which is naturally embedded in a product space of
image domain and the range of intensity values. For example, a planar
grey-value image $u:\mathbb{R}^2\supset\varOmega\to\mathbb{R}$ has the image
manifold $\{(\bm{x},u(\bm{x}))~|~\bm{x}\in\varOmega\}\subset\mathbb{R}^3$.
The dimension of this structure can be considered as a fractal-based texture
descriptor. This approach has been stated in \cite{Pentland-PAMI84}
where fractal dimension was put into relation with image roughness and the
roughness of physical structures depicted in the image.
Many works followed this approach, particularly in the 1980s and beginning
1990s when fractals were under particularly intensive investigation in
theoretical and applied mathematics. In \cite{Avadhanam-msc93}
several of these approaches are reviewed. An attempt to analyse
fractal dimension concepts for texture analysis is found in
\cite{Soille-JVCIR96}. The concept has also been transferred to
the analysis of 1D signals, see
\cite{Maragos-icassp91,Maragos-TSP93}.
During the last two decades the interest in fractal methods has somewhat
reduced but research in the field remains ongoing as can be seen from
more recent publications, see e.g.\ \cite{Pitsikalis-SC09} for signal
analysis, \cite{Cikalova-PMM11,Kuznetsov-TAFM01} for image
analysis with application in material science.
With regard to our analysis in Section~\ref{sec-tdana} that leads to
a relation between graph methods and fractals, it is worth mentioning that
already \cite{Torres-PR04} linked graph and fractal methods in image
analysis, albeit not considering texture but shape description.

\paragraph{Texture segmentation.}
The task of texture segmentation, i.e.\ decomposing an image into several
segments based on texture differences, has been studied for more than
forty years, see \cite{Rosenfeld-TC71,Sutton-TC72,
Zucker-CGIP76}. A great variety of different approaches to the
problem have been proposed since then. Many of these combine generic
segmentation approaches, that could also be implemented for merely
intensity-based segmentation, with sets of quantitative texture descriptors
that are used as inputs to the segmentation.
For example, \cite{Brox-IVC10,Paragios-JVCIR02,
Sagiv-TIP06,Sandberg-tr-cam02-39} are based on active contour
or active region models for segmentation, whereas
\cite{Georgescu-iccv03} is an example of a clustering-based method.
Nevertheless, texture segmentation continues to challenge researchers;
in particular, improvements on the side of texture descriptors are still
desirable.

Note that the task of texture segmentation involves a conflict: On one hand,
textures cannot be detected on single-pixel level, necessitating the
inclusion of neighbourhoods in texture descriptor computation. On the other
hand, the intended output of a segmentation is a unique assignment of each
pixel to a segment, which means to fix the segment boundaries at pixel
resolution. To allow sufficiently precise location of boundaries, texture
descriptors should therefore not use larger patches than necessary to
distinguish the textures present in an image.

The content of this paper is centred around a set of texture descriptors
that have been introduced in \cite{Welk-qgt14} based on graph
representations of local image structure. This model seems to be the
first that exploited graph models in discrimination of textures. Note
that even texture segmentation approaches in literature that use graph cuts
for the segmentation task use non-graph-based texture descriptors to bring the
texture information into the graph-cut formulation,
see e.g.\ \cite{Shi-PAMI00}.
Our texture segmentation approach that was already shortly demonstrated in
\cite{Welk-bookchapter15x} integrates instead graph-based texture
descriptors into a non-graph-based segmentation framework, compare
Section~\ref{sec-gac}.

\section{Graph-Entropy-Based Texture Descriptors}
\label{sec-gi}

Throughout the history of texture processing, quantitative texture
descriptors have played an important role. Assigning a tuple of
numerical values to a texture provides an interface to established
image processing algorithms that were originally designed to act
on intensities, and thereby to devise modular frameworks for image
processing tasks that involve texture information.

Following this modular approach, we will attack the texture segmentation
task by combining a set of texture descriptors with a well-established
image segmentation algorithm. In this section we will introduce the
texture descriptors whereas the following section will be devoted
to describing the segmentation method.

Given the variety of different textures that exist in natural images,
it cannot be expected that one single texture descriptor will be
suitable to discriminate arbitrary textures. Instead, it will be sensible
to come up with a set of descriptors that complement each other well
in distinguishing different kinds of textures. To keep the set of
descriptors at a manageable size, the individual descriptors should
nevertheless be able to discriminate substantial classes of textures.
On the other hand,
it will be useful both for theoretical analysis and for practical
computation if the set of descriptors is not entirely disparate but
based on some powerful common concept.

In \cite{Welk-qgt14} a family of texture descriptors was introduced
based on the application of several graph indices to
graphs representing local image information. In combining six
sets of graphs derived from an image, whose computation is based on common
principles, with a number of different but related graph indices,
this family of descriptors is indeed built on a common concepts.

The descriptors were evaluated in \cite{Welk-qgt14}
in a simple texture discrimination
task, and turned out to yield results competitive with
Haralick features \cite{Haralick-PIEEE79,Haralick-TSMC73},
a well-established concept in texture analysis.
In this comparison, graph indices based on entropy measures stood
out by their texture discrimination rate.

In the following, we recall the construction of
texture descriptors from \cite{Welk-qgt14}, focussing on a subset
of the descriptors discussed there.
The first step is the construction
of graphs from image patches. In the second step, graph indices are
computed from these graphs.

\subsection{Graph Construction}
\label{ssec-graphcons}

A discrete grey-scale image is given as an array of real intensity values
sampled at the nodes of a regular grid. The nodes are points
$(x_i,y_j)$ in the plane where $x_i=x_0+ih_x$, $y_j=y_0+jh_y$.
The spatial mesh sizes $h_x$ and $h_y$ are often assumed to be $1$ in
image processing, which we will do also here for simplicity.
Denoting the intensity values by $u_{i,j}$ and assuming that
$i\in\{0,\ldots,n_x\}$, $j\in\{0,\ldots,n_y\}$, the image is then
described as the array $\bm{u}=(u_{i,j})$.

The nodes of the grid, thus the pixels of the image, can naturally be
considered as vertices of a graph in which neighbouring pixels are
connected by edges. We will call this graph the \emph{pixel graph} $G_u$
of the image. Two common choices for what pixels are considered
as neighbours are based on $4$-neighbourhoods, in which pixel $(i,j)$ has
the two horizontal neighbours $(i\pm1,j)$ and the two vertical neighbours
$(i,j\pm1)$, or $8$-neighbourhoods, in which also the four diagonal
neighbours $(i\pm1,j\pm1)$ are included in the neighbourhood.
Whereas the $4$-neighbourhood setting leads to a somewhat simpler pixel
graph (particularly, it is planar), whereas $8$-neighbourhoods are
better suited to capture the geometry of the underlying (Euclidean)
plane. In this work, we will mostly use $8$-neighbourhoods.
See \cite[Sec.~1.5]{Lezoray-bookchapter12} for more variants of graphs
assigned to images.

We upgrade the pixel graph to an \emph{edge-weighted pixel graph} $G_w$ by
defining edge weights $w_{\bm{p},\bm{q}}$ for neighbouring
pixels $\bm{p},\bm{q}$ via
\begin{equation}
w_{\bm{p},\bm{q}} :=
\left(
\lVert\bm{p}-\bm{q}\rVert^2+\beta^2\lvert u_{\bm{p}}-u_{\bm{q}}\rvert^2
\right)^{1/2}\;,
\end{equation}
i.e.\ an $l_2$ sum of the spatial distance of grid nodes
$\lVert\bm{p}-\bm{q}\rVert$ (where $\lVert\,\cdot\rVert$ denotes the
Euclidean norm), and the contrast $\lvert u_{\bm{p}}-u_{\bm{q}}\rvert$
of their corresponding intensity values, weighted by a positive
\emph{contrast scale} $\beta$. This construction can of course
be generalised by replacing the Euclidean norm in the image plane, and
the $l_2$ sum by other norms. With various settings for these norms,
it has been used e.g.\ in \cite{Lerallut-ismm05,Lerallut-IVC07,
Welk-Aiep14,Welk-JMIV11} to construct larger
spatially adaptive neighbourhoods in images, so-called
\emph{morphological amoebas.}
See also \cite{Welk-qgt14} for a more detailed
description of the amoeba framework in a graph-based terminology.

All graphs that will enter the texture descriptor construction are
derived from the pixel graph or the edge-weighted pixel graph of the image.
First, given a pixel $\bm{p}$ and a radius $\varrho>0$, we define the
Euclidean patch graph $G_w^{\mathrm{E}}(\bm{p},\varrho)$ as the subgraph
of $G_w$ which includes all nodes $\bm{q}$ with Euclidean distance
$\lVert\bm{q}-\bm{p}\rVert\le\varrho$. In this graph, image information
is encoded solely in the edge weights.

Second, we define the adaptive patch graph $G_w^{\mathrm{A}}(\bm{p},\varrho)$
as the subgraph of $G_w$ which includes all nodes $\bm{q}$ for which
$G_w$ contains a path from $\bm{p}$ to $\bm{q}$ with total weight
less or equal $\varrho$.
In the terminology of
\cite{Lerallut-ismm05,Lerallut-IVC07,
Welk-Aiep14,Welk-JMIV11}, the node set of
$G_w^{\mathrm{A}}(\bm{p},\varrho)$ is a morphological amoeba
of amoeba radius $\varrho$ around $\bm{p}$, which we will
denote by $\mathcal{A}_{\varrho}(\bm{p})$.
Note that the graph $G_w^{\mathrm{A}}(\bm{p},\varrho)$ encodes image
information not only in its edge weights, but also in its node set
$\mathcal{A}_{\varrho}(\bm{p})$.

One obvious way to compute $\mathcal{A}_{\varrho}(\bm{p})$ is by Dijkstra's
shortest path algorithm \cite{Dijkstra-NUMA59} with $\bm{p}$ as
starting point. A natural by-product of this algorithm, which is not used
in amoeba-based image filtering as in \cite{Lerallut-ismm05} etc.,
is the Dijkstra search tree, which we denote as
$T_w^{\mathrm{A}}(\bm{p},\varrho)$. This is the third candidate graph for
our texture description. Image information is encoded in this graph in
three ways: in the edge weights, the node set, and the connectivity of
the tree.

Dropping the edge weights from $T_w^{\mathrm{A}}(\bm{p},\varrho)$, we
obtain an unweighted tree $T_u^{\mathrm{A}}(\bm{p},\varrho)$ which still
encodes image information in its node set and connectivity.
Finally, a Dijkstra search tree $T_w^{\mathrm{E}}((\bm{p},\varrho)$ and
its unweighted counterpart $T_u^{\mathrm{E}}((\bm{p},\varrho)$ can be
obtained by applying Dijkstra's shortest path algorithm within
the Euclidean patch graph $G_w^{\mathrm{E}}(\bm{p},\varrho)$.
Whereas $T_w^{\mathrm{E}}((\bm{p},\varrho)$ encodes image information
in the edge weights and connectivity, $T_u^{\mathrm{E}}((\bm{p},\varrho)$
does so only in the connectivity.

Applying these procedures to all pixels $\bm{p}=(i,j)$ of a discrete image
$\bm{u}$, we have therefore six collections of graphs
which represent different combinations of three cues to local image
information (edge weights, node sets, connectivity) and can therefore be
expected to be suitable for texture discrimination. In the following we
will drop the arguments $\bm{p}$, $\varrho$ and use simply
$G_w^{\mathrm{A}}$ etc.\ to refer to the collections of graphs.

\subsection{Entropy-Based Graph Indices}
\label{ssec-gi}

In order to turn the collections of graphs into quantitative texture
descriptors suitable for texture analysis tasks, the realm of graph
indices developed in quantitative graph theory lends itself as a powerful
tool.

In \cite{Welk-qgt14}, a selection of graph indices was
considered for this purpose, including on one hand distance-based
graph indices (the Wiener index \cite{Wiener-JACS47}, the
Harary index \cite{Plavsic-JMC93} and the Balaban index
\cite{Balaban-CPL82}) and on the other hand entropy-based indices
(Bonchev-Trinajsti{\'c} indices
\cite{Bonchev-JCC81,Bonchev-JCP77} and
Dehmer entropies \cite{Dehmer-AMC08}).

The so-obtained set of $42$ texture descriptors was evaluated in
\cite{Welk-qgt14} with respect to their discrimination power
and diversity. Using nine textures from a database \cite{vistex},
texture discrimination power was quantified
based on simple statistics (mean value and standard deviation) of the
descriptor values within single-texture patches, calibrating thresholds
for certain and uncertain discrimination of textures within the set
of textures. Diversity of descriptors was measured based on the
overlap in the sets of texture pairs discriminated by different descriptors.
Despite the somewhat ad-hoc character of the threshold calibration, the
study gives valuable hints for the selection of powerful subsets of the $42$
texture descriptors.

Among the descriptors being analysed, particularly the
entropy-based descriptors ranked medium to high regarding discrimination
power for the sample set of textures. For the present work, we focus therefore
on three entropy measures which we will recall in the following, namely
the Dehmer entropies $I_{f^V}$ and $I_{f^P}$ as well as
Bonchev and Trinajsti{\'c}'s mean information on distances
$\bar{I}_{\mathrm{D}}^{\mathrm{E}}$. The latter is restricted by
its construction to unweighted graphs, and is therefore used with
the unweighted Dijkstra trees
$T_u^{\mathrm{A}}$ and $T_u^{\mathrm{E}}$.
The Dehmer entropies can be combined with all six graph collections.

In \cite{Welk-qgt14}, the Dehmer entropies on the
patch graphs $G_w^{\mathrm{A}}$ and $G_w^{\mathrm{E}}$ achieved the highest
rates of certain discrimination of textures, and outperformed the
Haralick features included in the study. Some of the other descriptors
based on Dehmer entropies as well as the Bonchev-Trinajsti{\'c} information
measures
achieved middle ranks, combining medium rates of certain discrimination with
uncertain discrimination of almost all other texture pairs; thereby, they
were still comparable to Haralick features and distance-based graph
indices.

\subsubsection{Shannon's Entropy}

The measures considered here are based on
Shannon's entropy \cite{Shannon-BSTJ48}
\begin{equation}
H(p) = -\sum_{i=1}^n p_i\,\operatorname{ld} p_i
\label{entropy}
\end{equation}
that measures the information content of a discrete probability measure
$p:\{1,\ldots,n\}\to\mathbb{R}^+_0$, $i\mapsto p_i$, $\sum_{i=1}^np_i=1$.
(Note that for $p_i=0$ one has to set in \eqref{entropy}
$p_i\,\operatorname{ld} p_i=0$ by limit.)

Following \cite{Dehmer-AMC08}, a discrete probability measure can be
assigned to an arbitrary nonnegative-valued function
$f:\{1,\ldots,n\}\to\mathbb{R}^+_0$, an \emph{information functional},
via
\begin{equation}
p_i:=\frac{f_i}{\sum_{j=1}^nf_j}\;.
\label{inffun-f2p}
\end{equation}
An entropy measure on an arbitrary information functional $f$ is then obtained
by applying \eqref{entropy} to \eqref{inffun-f2p}.

\subsubsection{Bonchev and Trinajsti{\'c}'s Mean Information on Distances}

Introduced in \cite{Bonchev-JCP77} and further investigated
in \cite{Bonchev-JCC81}, the \emph{mean information on distances}
is the entropy measure resulting from an information functional on the
path lengths in a graph. Let a graph $G$ with $n$ vertices $v_1,\ldots,v_n$
be given, and let $d(v_i,v_j)$ denote the length of a shortest path
from $v_i$ to $v_j$ in $G$ (unweighted, i.e.\ each edge counting $1$).
Let $D(G):=\max_{i,j}d(v_i,v_j)$ be the diameter of $G$.
For each $d\in\{1,\ldots,D(G)\}$, let
\begin{equation}
k_d := \# \{ (i,j)~|~1\le i<j\le n,~d(i,j)=d\}\;.
\end{equation}
The mean information on distances then is the entropy measure based on the
information functional $k_d$, i.e.\
\begin{equation}
\bar{I}_{\mathrm{D}}^{\mathrm{E}}(G) = 
-\sum_{d=1}^{D(G)}\frac{k_d}{\binom{n}{2}}
\,\operatorname{ld}\frac{k_d}{\binom{n}{2}}\;.
\end{equation}
Let us shortly mention that \cite{Bonchev-JCP77} also introduces
the \emph{mean information on realised distances}
$\bar{I}_{\mathrm{D}}^{\mathrm{W}}(G)$, which we will not further
consider here.
As an entropy measure, $\bar{I}_{\mathrm{D}}^{\mathrm{W}}(G)$ can be derived
from the information functional
$d(v_i,v_j)$ on the set of all vertex pairs $(v_i,v_j)$, $1\le i<j\le n$,
of $G$. As pointed out in \cite{Welk-qgt14},
$\bar{I}_{\mathrm{D}}^{\mathrm{W}}(G)$ can be generalised straightforwardly to
edge-weighted graphs by measuring distances $d(v_i,v_j)$ with edge weights,
but a similar generalisation of $\bar{I}_{\mathrm{D}}^{\mathrm{E}}(G)$ would
be degenerated because in generic cases all edge-weighted distances in a graph
will be distinct, leading to $k_d=1$ for all realised values $d$. Therefore
we will use the mean information on distances
$\bar{I}_{\mathrm{D}}^{\mathrm{E}}(G)$ only with the unweighted graphs
$T_u^{\mathrm{E}}$ and $T_u^{\mathrm{A}}$.

\subsubsection{Dehmer Entropies}

The two entropy measures $I_{f^V}(G)$ and $I_{f^P}(G)$ for unweighted
graphs $G$ were introduced in \cite{Dehmer-AMC08}. Their high
discriminative power for large sets of graphs was impressively
demonstrated in \cite{Dehmer-PLO12a}. Both measures rely
on information functionals on the vertex set $\{v_1,\ldots,v_n\}$
of $G$ whose construction involves spheres $S_d(v_i)$ of varying radius $d$
around $v_i$. Note that the sphere $S_d(v_i)$ in $G$ is the set of vertices
$v_j$ with $d(v_i,v_j)\le d$.

For $I_{f^V}$ the information functional $f^V$ on vertices
$v_i$ of an unweighted graph $G$ is defined as \cite{Dehmer-AMC08}
\begin{equation}
f^V(v_i) := \exp\left(\sum_{d=1}^{D(G)}c_d s_d(j)\right)\;,
\label{fV-Dehmer-e}
\end{equation}
where
\begin{equation}
s_d(j):=\# S_d(v_j)
\label{sd}
\end{equation}
is the cardinality of the $d$-sphere around $v_j$,
with positive parameters $c_1,\ldots,c_{D(G)}$.
(Note that \cite{Dehmer-AMC08} used a general exponential with
base $\alpha$. For the purpose of the present paper, however, this
additional parameter is easily eliminated
by multiplying the coefficients $c_i$ with $\ln\alpha$.)

For $I_{f^P}$ the information functional $f^P$ relies on the quantities
\begin{equation}
l_d(i):=\sum_{j:v_j\in S_d(v_i)}d(v_i,v_j)\;,
\label{ld}
\end{equation}
i.e.\ $l_d(i)$ is the sum of distances
from $v_i$ to all points in its $d$-sphere. With similar parameters
$c_1,\ldots,c_{D(G)}$ as before, one defines then
\begin{equation}
f^P(v_i):=\exp\left(\sum_{d=1}^{D(G)}c_d l_d(j)\right)\;.
\label{fP-Dehmer-e}
\end{equation}

As pointed out in \cite{Welk-qgt14}, both information functionals,
and thus the resulting entropy measures $I_{f^V}$, $I_{f^P}$, can be
adapted to edge-weighted graphs $G$ via
\begin{align}
f^V(v_i) &=\exp\left(\sum_{j=1}^{n}C(d(v_i,v_j))\right)\;,
\label{fV-weighted}
\\
f^P(v_i) &=\exp\left(\sum_{j=1}^{n}C(d(v_i,v_j)) d(v_i,v_j)\right)\;,
\label{fP-weighted}
\end{align}
where distances $d(v_i,v_j)$ are now measured using the edge weights,
and $C:[0,D(G)]\to\mathbb{R}^+_0$ is a decreasing function interpolating a
reverse partial sum series of the original $c_d$ coefficients.

Further following \cite{Welk-qgt14},
we focus on the specific choice
\begin{align}
c_d&= q^d\;,\quad q\in(0,1)
\label{expws}
\end{align}
(an instance of
the \emph{exponential weighting scheme} from \cite{Dehmer-PLO12a})
and obtain accordingly $C(d)=M q^d$ with a positive constant $M$,
which yields
\begin{align}
f^V(v_i) &= \exp\left(M\sum_{j=1}^n q^{d(v_i,v_j)}\right)\;,
\label{fV-weighted-expws}
\\
f^P(v_i) &= \exp\left(M\sum_{j=1}^n q^{d(v_i,v_j)}d(v_i,v_j)\right) \;,
\label{fP-weighted-expws}
\end{align}
with a positive constant $M$,
as the final form of the information functionals for our construction of
texture descriptors.

\section{Geodesic Active Contours}
\label{sec-gac}

We use for our experiments a well-established segmentation method
based on partial differential equations (PDE). Introduced in
\cite{Caselles-iccv95,Kichenassamy-iccv95},
\emph{geodesic active contours} (GAC) perform a contrast-based segmentation
of a (grey-scale) input image $f$.

Of course, other contrast-based segmentation methods could be
chosen, including clustering
\cite{Comaniciu-cvpr97,Lucchese-cbaivl99}
or graph-cut methods
\cite{Boykov-PAMI01,Ishikawa-bookchapter12}.
Advantages or disadvantages of these methods in connection with
graph-entropy-based texture descriptors may be studied in future work.
For the time being we focus on the texture descriptors themselves, thus
it matters to use just one well-established standard method.

\subsection{Basic GAC Evolution for Greyscale Images}
\label{ssec-gac-std}

From the input image $f$, a
Gaussian-smoothed image $f_\sigma:=G_\sigma*f$ is computed, where
$G_\sigma$ is a Gaussian kernel of standard deviation $\sigma$.
From $f_\sigma$, one computes an edge map $g(\lvert\bm{\nabla}f_\sigma\rvert)$
with the help of a decreasing and bounded function
$g:\mathbb{R}^+_0\to\mathbb{R}^+_0$ with $\lim_{s\to\infty}g(s)=0$.
A popular choice for $g$ is
\begin{equation}
g(s)=\frac1{1+s^2/\lambda^2}
\label{peronamalikg}
\end{equation}
which has originally been introduced by Perona and Malik
\cite{Perona-PAMI90} as a diffusivity function for nonlinear
diffusion filtering of images. Herein, $\lambda>0$ is a contrast parameter
that acts as a threshold distinguishing high gradients (indicating
probable edges) from small ones.

In addition to the input image, GAC require an initial contour $C_0$
(a regular closed curve) specified
e.g.\ by user input. This contour is embedded into a \emph{level set
function} $u_0$, i.e.\ $u_0$ is a sufficiently smooth function in the image
plane whose zero level-set (set of all points $(x,y)$ in the image plane
for which $u_0(x,y)=0$) is the given contour. For example, $u_0$ can be
introduced as a signed distance function: $u_0(x,y)$ is
zero if $(x,y)$ lies on $C_0$; it is minus the distance of $(x,y)$ to $C_0$
if $(x,y)$ lies in the region enclosed by $C_0$, and plus the same distance
if $(x,y)$ lies in the outer region.

One takes then $u_0$ as initial condition at time $t=0$ for the parabolic
PDE
\begin{equation}
u_t = \lvert\bm{\nabla}u\rvert\operatorname{div}\left(
g(\lvert\bm{\nabla}f_\sigma\rvert)\frac{\bm{\nabla}u}{\lvert\bm{\nabla}u\rvert}
\right)
\label{gac}
\end{equation}
for a time-dependent level-set function $u(x,y,t)$. At each time $t\ge0$, an
evolved contour can be extracted from $u(\,\cdot\,,\,\cdot\,,t)$ as zero
level-set. For suitable input images and initialisations and with appropriate
parameters, the contours lock in at a steady state that provides a
contrast-based segmentation.

To understand equation \eqref{gac} one can compare it to
the curvature motion
equation $u_t=\operatorname{div}(\bm{\nabla}u/\lvert\bm{\nabla}u\rvert)$
that would evolve all level curves of $u$ by an inward movement proportional
to their curvature. In \eqref{gac}, this inward movement of level
curves is modulated by the edge map $g(\lvert\bm{\nabla}f_\sigma\rvert)$,
which slows down the curve displacement at high-contrast locations, such that
contours stick there.

The name \emph{geodesic active contours} is due to the fact that
the contour evolution associated to \eqref{gac} can be understood as
gradient descent for the curve length of the contour in an image-adaptive
metric (a Riemannian metric whose metric tensor is
$g(\lvert\bm{\nabla}f_\sigma\rvert)$ times the unit matrix), thus yielding
a final contour that is a geodesic with respect to this metric.

\subsection{Force Terms}
\label{ssec-gac-force}

In its pure form \eqref{gac}, geodesic active contours require the
initial contour (at least most of it) to be placed outside the object to
be segmented. In some situations, however, it is easier to specify an
initial contour inside an object, particularly if the intensities within
the object are fairly homogeneous but many spurious edges irritating the
segmentation exist in the background region.

Moreover, despite being able to handle also topology changes
such as a splitting from one to several level curves encircling distinct
objects to some extent, it has limitations when the topology of the
segmentation becomes too complex.

As a remedy to both difficulties, one can modify \eqref{gac} by
adding a \emph{force term} $\nu g(\lvert\bm{\nabla}f_\sigma\rvert)
\lvert\bm{\nabla}f_\sigma\rvert$ to its right-hand side.
Adding a force term was proposed first in
\cite{Cohen-CVGIPIU91} (by the name of \emph{balloon force})
whereas the specific form of the force term weighted with $g$ was
proposed in \cite{Caselles-iccv95,Kichenassamy-iccv95,
Malladi-PAMI95}. Depending on the sign of $\nu$, this force
exercises an inward ($\nu>0$) or outward ($\nu<0$) pressure on the
contour, which (i) speeds up the curve evolution, (ii) supports the
handling of complex segmentation topologies, and (iii) enables for
$\nu<0$ also segmentation of objects from initial contours placed inside.
{\sloppy\par}

The modified GAC evolution with force term,
\begin{equation}
u_t = \lvert\bm{\nabla}u\rvert\left(\operatorname{div}\left(
g(\lvert\bm{\nabla}f_\sigma\rvert)\frac{\bm{\nabla}u}{\lvert\bm{\nabla}u\rvert}
\right)
+\nu\,g(\lvert\bm{\nabla}f_\sigma\rvert)\right)
\label{gac-force}
\end{equation}
will be our segmentation
method when performing texture segmentation based on only one
quantitative texture descriptor. In this case, the texture descriptor
will be used as input image $f$ from which the edge map $g$ is
computed.

\subsection{Multi-Channel Images}
\label{ssec-gac-mc}

It is straightforward to extend the GAC method, including its modified
version with force term to multi-channel input images $\bm{f}$ where each
location $(x,y)$ in the image plane is assigned an $r$-tuple
$(f_1(x,y),\ldots,f_r(x,y))$ of intensities. A common case, with $r=3$,
are RGB colour images.

In fact, equations \eqref{gac} and \eqref{gac-force} incur
almost no change as even for multi-channel input images, one computes the
evolution of a simple real-valued level-set function $u$.
What is changed is the computation of the edge map $g$: Instead of
the gradient norm $\lvert \bm{\nabla}f_\sigma\rvert$ one uses the Frobenius
norm $\lVert \mathbf{D}\bm{f}_\sigma\rVert$ of the Jacobian
$\mathbf{D}\bm{f}_\sigma$ where $\bm{f}_\sigma$ is the Gaussian-smoothed
input image, $\bm{f}_\sigma = (f_{\sigma;1},\ldots,f_{\sigma;r})$ with
$f_{\sigma;i}=G_\sigma*f_i$, yielding
$g(\lVert \mathbf{D}\bm{f}_\sigma\rVert)$ as edge map.

Equation \eqref{gac-force} with this edge map will be our
segmentation method when performing texture segmentation with
multiple texture descriptors. The input image $\bm{f}$ will have the
individual texture descriptors as channels. To weight the influence
of texture descriptors, the channels may be multiplied by scalar
factors.

\subsection{Remarks on Numerics}
\label{ssec-gac-num}

For numerical computation we rewrite PDE \eqref{gac-force} as
\begin{equation}
u_t = g\,\lvert\bm{\nabla}u\rvert
\operatorname{div}\left(\frac{\bm{\nabla}u}{\lvert\bm{\nabla}u\rvert}
\right)+\langle\bm{\nabla}g,\bm{\nabla}u\rangle
+\nu\,g\,\lvert\bm{\nabla}u\rvert
\end{equation}
(where we have omitted the argument of $g$ which is a fixed input
function to the PDE anyway).

Following established practice,
we use then an explicit (Euler forward) numerical scheme where the
right-hand side is spatially discretised as follows.
The first term, $g\,\lvert\bm{\nabla}u\rvert
\operatorname{div}(\bm{\nabla}u/\lvert\bm{\nabla}u\rvert)$,
is discretised using central differences.
For the second term, $\langle\bm{\nabla}g,\bm{\nabla}u\rangle$,
an upwind discretisation \cite{Courant-CPAM52,Rouy-SINUM92}
is used in which the upwind direction for $u$ is determined based on
the central-difference approximations of $\bm{\nabla}g$.
The third term, $\nu\,g\,\lvert\bm{\nabla}u\rvert$, is discretised with
an upwind discretisation, too. Here, the upwind direction depends on the
components of $\bm{\nabla}u$ and the sign of $\nu$.{\sloppy\par}

Although a detailed stability analysis for this widely used type
of explicit scheme for the GAC equation seems to be missing, the scheme
works for time step sizes $\tau$ up to ca.\ $0.25$ (for spatial
mesh sizes of $h_x=h_y=1$) for $\nu=0$, which needs to be reduced somewhat
for non-zero $\nu$. In our experiments in
Section~\ref{sec-segex} we use consistently $\tau=0.1$.

For the level-set function $u$, we use the signed distance function of
the initial contour as initialisation. Since during the evolution the
shape of the level-set function changes, creating steeper ascents in
some regions but flattening slopes elsewhere, we re-initialise $u$ to
the signed distance function of its current zero level set every $100$
iterations.

\section{Texture Segmentation Experiments}
\label{sec-segex}

In this section, we present experiments on two synthetic and one
real-world test image that demonstrate that graph-entropy-based
texture descriptors can be used for texture-based segmentation.
An experiment similar to our second synthetic example was already
presented in \cite{Welk-bookchapter15x}.

\subsection{First Synthetic Example}
\label{ssec-segex-1}

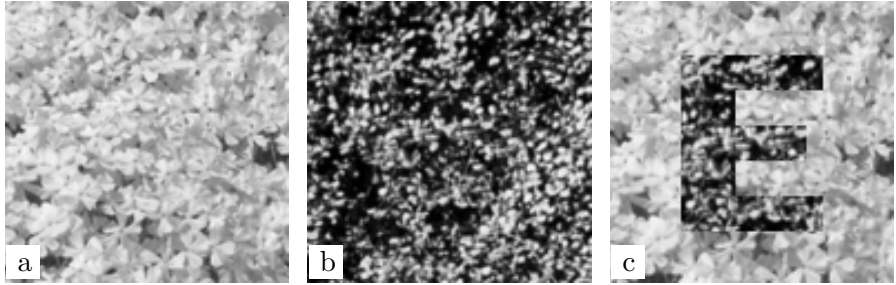
\begin{figure}[t!]
\unitlength0.001\textwidth
\begin{picture}(1000, 238)
\put( 127,   0){\includegraphics[width=238\unitlength]{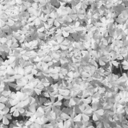}}
\put( 128,  10){\colorbox{white}{\rule{0pt}{.6em}\hbox to.6em{\kern.1em\smash{a}}}}
\put( 381,   0){\includegraphics[width=238\unitlength]{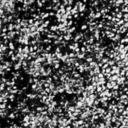}}
\put( 382,  10){\colorbox{white}{\rule{0pt}{.6em}\hbox to.6em{\kern.1em\smash{b}}}}
\put( 635,   0){\includegraphics[width=238\unitlength]{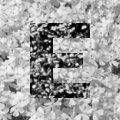}}
\put( 636,  10){\colorbox{white}{\rule{0pt}{.6em}\hbox to.6em{\kern.1em\smash{c}}}}
\end{picture}
\caption{\textbf{Left to right:}
\textbf{(a)} Texture patch \emph{flowers}, $128\times128$ pixels. --
\textbf{(b)} Texture patch \emph{leaves}, same size. --
\textbf{(c)} Test image composed from (a) and (b), $120\times120$ pixels. --
Both texture patches are converted to greyscale, downscaled and clipped
from the \emph{VisTex} database \cite{vistex}.
\copyright 1995 Massachusetts Institute of Technology. Developed by
Rosalind Picard, Chris Graczyk, Steve Mann, Josh Wachman, Len Picard,
and Lee Campbell at the Media Laboratory, MIT, Cambridge, Massachusetts.
Under general permission for scholarly use.
}
\label{fi-e3}
\end{figure}

In our first example we use a synthetic image, shown in
Figure~\ref{fi-e3}(c), which is composed from two textures,
see Figure~\ref{fi-e3}(a) an (b), with a simple shape (the
letter `E') switching between the two. Note that the two textures
were also among the nine textures studied in \cite{Welk-qgt14}
for the texture discrimination task.
With its use of real-world textures, this synthetic example mimicks
a realistic segmentation task. Its synthetic construction warrants
at the same time a ground truth to compare segmentation results with.

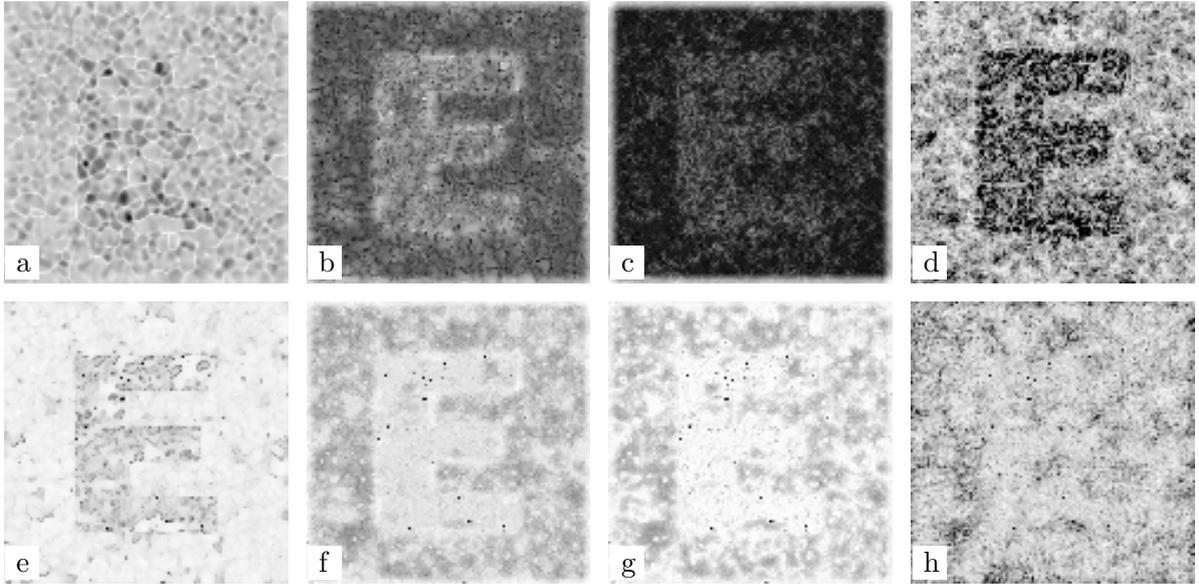
\begin{figure}[t!]
\unitlength0.001\textwidth
\begin{picture}(1000, 492)
\put(   0, 254){\includegraphics[width=238\unitlength]{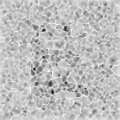}}
\put(   1, 264){\colorbox{white}{\rule{0pt}{.6em}\hbox to.6em{\kern.1em\smash{a}}}}
\put( 254, 254){\includegraphics[width=238\unitlength]{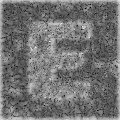}}
\put( 255, 264){\colorbox{white}{\rule{0pt}{.6em}\hbox to.6em{\kern.1em\smash{b}}}}
\put( 508, 254){\includegraphics[width=238\unitlength]{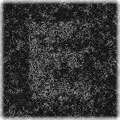}}
\put( 509, 264){\colorbox{white}{\rule{0pt}{.6em}\hbox to.6em{\kern.1em\smash{c}}}}
\put( 762, 254){\includegraphics[width=238\unitlength]{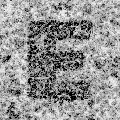}}
\put( 763, 264){\colorbox{white}{\rule{0pt}{.6em}\hbox to.6em{\kern.1em\smash{d}}}}
\put(   0,   0){\includegraphics[width=238\unitlength]{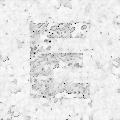}}
\put(   1,  10){\colorbox{white}{\rule{0pt}{.6em}\hbox to.6em{\kern.1em\smash{e}}}}
\put( 254,   0){\includegraphics[width=238\unitlength]{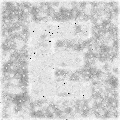}}
\put( 255,  10){\colorbox{white}{\rule{0pt}{.6em}\hbox to.6em{\kern.1em\smash{f}}}}
\put( 508,   0){\includegraphics[width=238\unitlength]{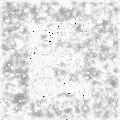}}
\put( 509,  10){\colorbox{white}{\rule{0pt}{.6em}\hbox to.6em{\kern.1em\smash{g}}}}
\put( 762,   0){\includegraphics[width=238\unitlength]{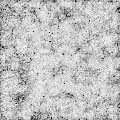}}
\put( 763,  10){\colorbox{white}{\rule{0pt}{.6em}\hbox to.6em{\kern.1em\smash{h}}}}
\end{picture}
\caption{Graph-entropy-based texture descriptors applied to the
test image from Figure~\ref{fi-e3}(c) (values rescaled to
$[0,255]$).
Patch radii were fixed to $\varrho=5$, contrast scale to $\beta=0.1$,
weighting parameter to $q=0.1$.
\textbf{Top row, left to right:}
\textbf{(a)} $I_{f^V}(G^{\mathrm{E}}_{w})$. --
\textbf{(b)} $I_{f^V}(T^{\mathrm{E}}_{w})$. --
\textbf{(c)} $I_{f^V}(T^{\mathrm{E}}_{u})$. --
\textbf{(d)} $I_{f^P}(T^{\mathrm{E}}_{w})$. --
\textbf{Bottom row, left to right:}
\textbf{(e)} $I_{f^V}(G^{\mathrm{A}}_{w})$. --
\textbf{(f)} $I_{f^V}(T^{\mathrm{A}}_{w})$. --
\textbf{(g)} $I_{f^V}(T^{\mathrm{A}}_{u})$. --
\textbf{(h)} $I_{f^P}(T^{\mathrm{A}}_{w})$.
}
\label{fi-e3-gi}
\end{figure}

Figure~\ref{fi-e3-gi} shows the results of eight graph-entropy-based
texture descriptors for the test image. In particular, the combination of
the Dehmer entropy $I_{f^V}$ with all six graph variants from
Section~\ref{ssec-graphcons} is shown as well as $I_{f^P}$ on the
weighted Dijkstra trees in non-adaptive and adaptive patches. Patch radii
were fixed to $\varrho=5$ for both non-adaptive and adaptive patches,
whereas the contrast scale was chosen as $\beta=0.1$. These parameter
settings have already been used in \cite{Welk-qgt14}; they
are based on values that work across various test images in the
context of morphological amoeba image filtering. Further investigation
of variation of these parameters is left for future work.

Visual inspection of Figure~\ref{fi-e3-gi} indicates that for this
specific textured image, the entropy measure $I_{f^V}$ separates the
two textures well in combination in particular with the weighted Dijkstra
tree settings, in both adaptive and non-adaptive patches, see frames
(b) and (f). The other $I_{f^V}$ results in frames (c), (e) and (g) show
insufficient contrast along some parts of the contour of the letter `E'.
The index $I_{f^V}(G_w^{\mathrm{E}})$ in frame (a), which was identified
in \cite{Welk-qgt14} as a descriptor with high texture discrimination
power, does not distinguish these two textures clearly but creates massive
over-segmentation within each of them. In a sense, this over-segmentation
is the downside of the high texture discrimination power of the descriptor.
Note, however, that also other $G_w^{\mathrm{E}}$-based descriptors tend
to this kind of over-segmentation.

Regarding the $I_{f^P}$ index, Figure~\ref{fi-e3-gi} (d) and (h),
there is a huge difference between the adaptive and non-adaptive patch
setting. Distinction of the two textures is much better when
using non-adaptive patches.

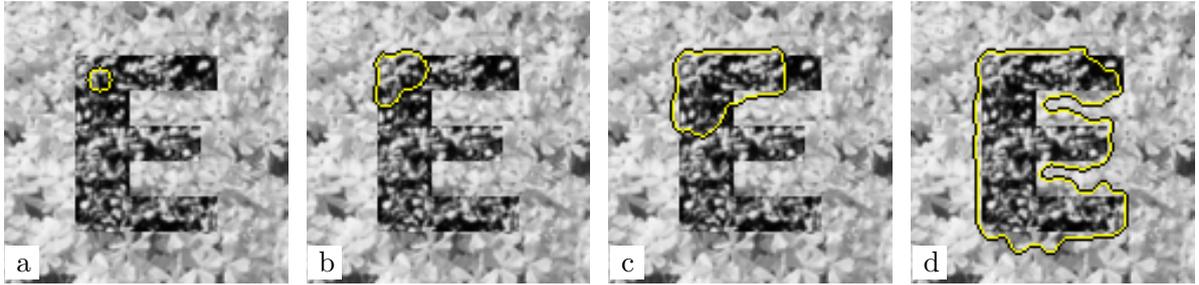
\begin{figure}[t!]
\unitlength0.001\textwidth
\begin{picture}(1000, 238)
\put(   0,   0){\includegraphics[width=238\unitlength]{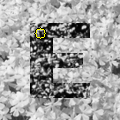}}
\put(   1,  10){\colorbox{white}{\rule{0pt}{.6em}\hbox to.6em{\kern.1em\smash{a}}}}
\put( 254,   0){\includegraphics[width=238\unitlength]{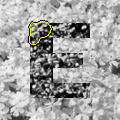}}
\put( 255,  10){\colorbox{white}{\rule{0pt}{.6em}\hbox to.6em{\kern.1em\smash{b}}}}
\put( 508,   0){\includegraphics[width=238\unitlength]{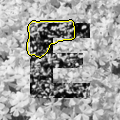}}
\put( 509,  10){\colorbox{white}{\rule{0pt}{.6em}\hbox to.6em{\kern.1em\smash{c}}}}
\put( 762,   0){\includegraphics[width=238\unitlength]{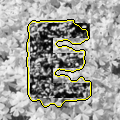}}
\put( 763,  10){\colorbox{white}{\rule{0pt}{.6em}\hbox to.6em{\kern.1em\smash{d}}}}
\end{picture}
\caption{%
Geodesic active contour segmentation of the image shown in
Figure~\ref{fi-e3}(c).
The edge map is computed from the graph-entropy-based texture descriptor
$I_{f^V}(T^{\mathrm{A}}_{w})$
from Figure~\ref{fi-e3-gi}(f) using pre-smoothing with $\sigma=2$,
Perona-Malik edge-stopping function \eqref{peronamalikg}
with $\lambda=0.1$, with expansion force $\nu=-1$.
\textbf{Left to right:}
\textbf{(a)} Initial contour ($t=0$). --
\textbf{(b)} $t=40$. --
\textbf{(c)} $t=160$. --
\textbf{(d)} $t=490$ (steady state).
}
\label{fi-e3-gac}
\end{figure}

Finally, we show in Figure~\ref{fi-e3-gac} geodesic active contour
segmentation of the test image with the descriptor
$I_{f^V}(T^{\mathrm{A}}_{w})$. We start from an initial contour inside the
`E' shape, see Figure~\ref{fi-e3-gac}(a),
and use an expansion force ($\nu=-1$) to drive the contour
evolution in an outward direction. Frames (b) and (c) show two intermediary
stages of the evolution, where it is evident that the contour starts to
align with the boundary between the two textures. Frame (d) shows
the steady state reached after $4\,900$ iterations ($t=490$).
Here, the overall shape of the letter `E'
is reasonably approximated, with deviations coming from small-scale
texture details.

Precision of the segmentation could be increased slightly by combining
more than one texture descriptor. We do not follow this direction at this
point.

\subsection{Second Synthetic Example}
\label{ssec-segex-2}

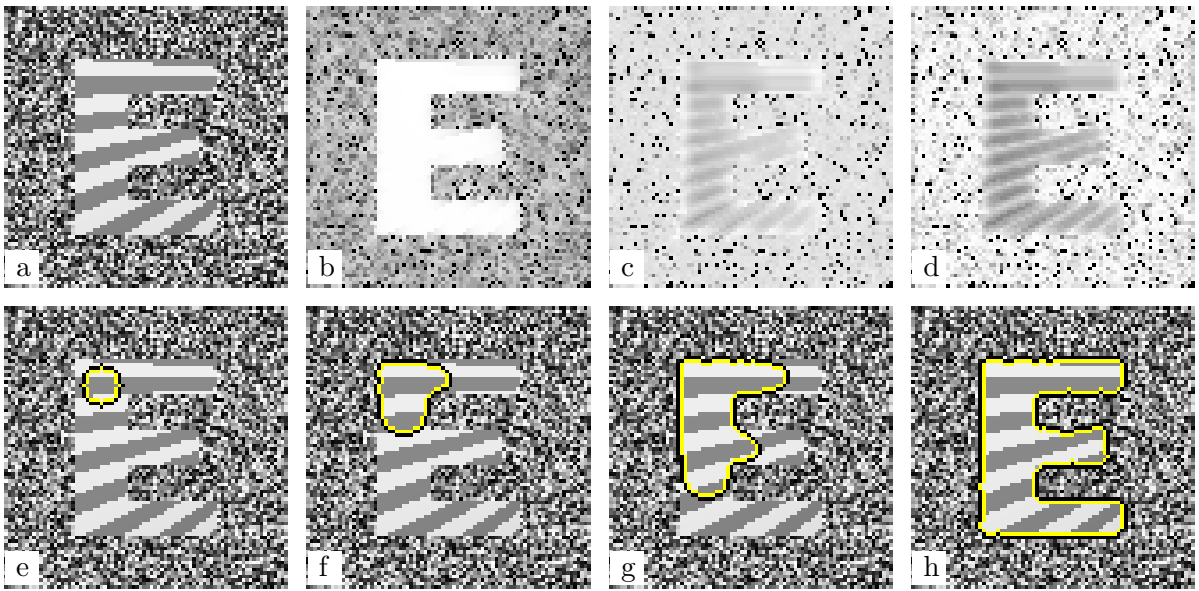
\begin{figure}[t!]
\unitlength0.001\textwidth
\begin{picture}(1000, 492)
\put(   0, 254){\includegraphics[width=238\unitlength]{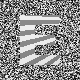}}
\put(   1, 264){\colorbox{white}{\rule{0pt}{.6em}\hbox to.6em{\kern.1em\smash{a}}}}
\put( 254, 254){\includegraphics[width=238\unitlength]{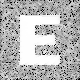}}
\put( 255, 264){\colorbox{white}{\rule{0pt}{.6em}\hbox to.6em{\kern.1em\smash{b}}}}
\put( 508, 254){\includegraphics[width=238\unitlength]{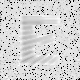}}
\put( 509, 264){\colorbox{white}{\rule{0pt}{.6em}\hbox to.6em{\kern.1em\smash{c}}}}
\put( 762, 254){\includegraphics[width=238\unitlength]{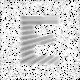}}
\put( 763, 264){\colorbox{white}{\rule{0pt}{.6em}\hbox to.6em{\kern.1em\smash{d}}}}
\put(   0,   0){\includegraphics[width=238\unitlength]{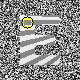}}
\put(   1,  10){\colorbox{white}{\rule{0pt}{.6em}\hbox to.6em{\kern.1em\smash{e}}}}
\put( 254,   0){\includegraphics[width=238\unitlength]{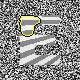}}
\put( 255,  10){\colorbox{white}{\rule{0pt}{.6em}\hbox to.6em{\kern.1em\smash{f}}}}
\put( 508,   0){\includegraphics[width=238\unitlength]{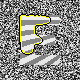}}
\put( 509,  10){\colorbox{white}{\rule{0pt}{.6em}\hbox to.6em{\kern.1em\smash{g}}}}
\put( 762,   0){\includegraphics[width=238\unitlength]{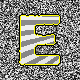}}
\put( 763,  10){\colorbox{white}{\rule{0pt}{.6em}\hbox to.6em{\kern.1em\smash{h}}}}
\end{picture}
\caption{%
Texture segmentation of a synthetic image.
\textbf{Top row, left to right:}
\textbf{(a)} Test image ($80\times80$ pixels) showing a stripe-textured
shape in front of a noise background. --
\textbf{(b--d)} Texture descriptors based on graph entropies applied
in adaptive patches, $\varrho=5$, $\beta=0.1$, $q=0.1$;
values rescaled to $[0,255]$.
\textbf{(b)} $I_{f^V}(G^{\mathrm{A}}_{w})$. --
\textbf{(c)} $I_{f^V}(T^{\mathrm{A}}_{w})$. --
\textbf{(d)} $I_{f^V}(T^{\mathrm{A}}_{u})$. --
\textbf{Bottom row:} Geodesic active contour segmentation of (a) using
the texture descriptor ($I_{f^V}$ on $G^{\mathrm{A}}_{w}$) from (b),
same parameters as in Figure~\ref{fi-e3-gi} except for $\sigma=1$.
\textbf{Left to right:}
\textbf{(e)} Initial contour ($t=0$). --
\textbf{(f)} $t=10$. --
\textbf{(g)} $t=30$. --
\textbf{(h)} $t=110$ (steady state).
}
\label{fi-e-stripes}
\end{figure}

In our second experiment, Figure~\ref{fi-e-stripes}, we use again a
synthetic test image where foreground and background segments are defined
using the `E' letter shape such that again the desired segmentation
result is known as a ground truth.
Also in this image we combine two realistic textures which can be seen
as a simplified version of the foreground and background textures
of the real-world test image, Figure~\ref{fi-zebra}(a),
used in the next section.
This time, the foreground is filled with a
stripe pattern whereas the background is noise with uniform distribution in
the intensity range $[0,255]$, see Figure~\ref{fi-e-stripes}(a).
In frames (b)--(d) of Figure~\ref{fi-e-stripes}
we show the texture descriptors based on $I_{f^V}$ with
the three graph settings in adaptive patches, using again $\varrho=5$ and
$\beta=0.1$. The descriptor $I_{f^V}(T_w^{\mathrm{A}})$ that was used for
the segmentation in Section~\ref{ssec-segex-1} visually does not
distinguish foreground from background satisfactorily here, whereas
$I_{f^V}(G_w^{\mathrm{A}})$ that provided no clear distinction of the
two textures in Section~\ref{ssec-segex-1} clearly stands out here.
This underlines the necessity of considering multiple descriptors which
complement each other in distinguishing textures.

Our GAC segmentation of the test image shown in frames (e)--(h) is based on
the texture descriptor $I_{f^V}(G_w^{\mathrm{A}})$ and quickly converges to
a fairly good approximation of the segment boundary.

\subsection{Real-World Example}
\label{ssec-segex-3}

\begin{figure}[t!]
\unitlength0.001\textwidth
\begin{picture}(1000, 758)
\put(   0, 516){\includegraphics[width=322\unitlength]{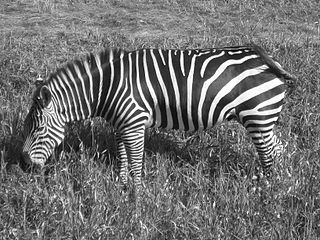}}
\put(   1, 526){\colorbox{white}{\rule{0pt}{.6em}\hbox to.6em{\kern.1em\smash{a}}}}
\put( 339, 516){\includegraphics[width=322\unitlength]{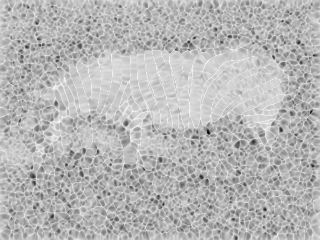}}
\put( 340, 526){\colorbox{white}{\rule{0pt}{.6em}\hbox to.6em{\kern.1em\smash{b}}}}
\put( 678, 516){\includegraphics[width=322\unitlength]{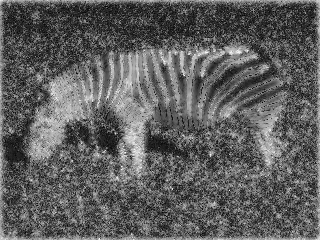}}
\put( 680, 526){\colorbox{white}{\rule{0pt}{.6em}\hbox to.6em{\kern.1em\smash{c}}}}
\put(   0, 258){\includegraphics[width=322\unitlength]{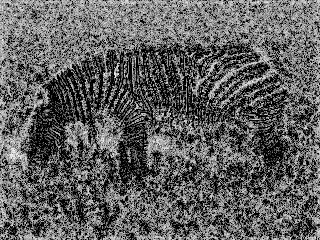}}
\put(   1, 268){\colorbox{white}{\rule{0pt}{.6em}\hbox to.6em{\kern.1em\smash{d}}}}
\put( 339, 258){\includegraphics[width=322\unitlength]{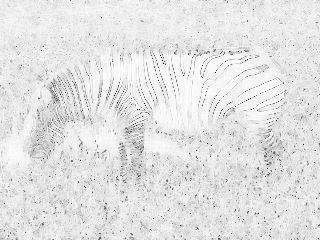}}
\put( 340, 268){\colorbox{white}{\rule{0pt}{.6em}\hbox to.6em{\kern.1em\smash{e}}}}
\put( 678, 258){\includegraphics[width=322\unitlength]{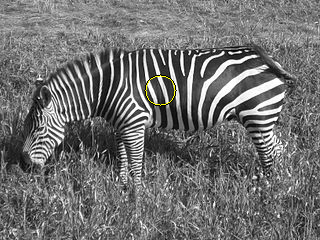}}
\put( 680, 268){\colorbox{white}{\rule{0pt}{.6em}\hbox to.6em{\kern.1em\smash{f}}}}
\put(   0,   0){\includegraphics[width=322\unitlength]{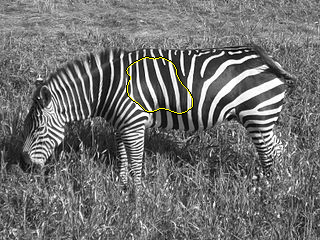}}
\put(   1,  10){\colorbox{white}{\rule{0pt}{.6em}\hbox to.6em{\kern.1em\smash{g}}}}
\put( 339,   0){\includegraphics[width=322\unitlength]{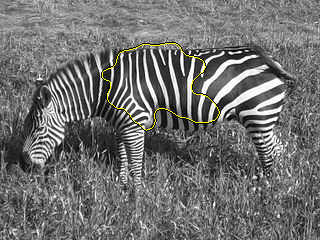}}
\put( 340,  10){\colorbox{white}{\rule{0pt}{.6em}\hbox to.6em{\kern.1em\smash{h}}}}
\put( 678,   0){\includegraphics[width=322\unitlength]{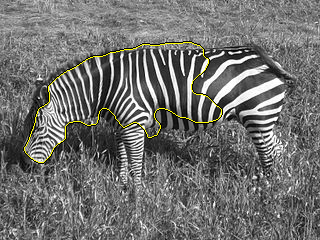}}
\put( 680,  10){\colorbox{white}{\rule{0pt}{.6em}\hbox to.6em{\kern.1em\smash{i}}}}
\end{picture}
\caption{
Texture segmentation of a real-world image.
\textbf{Top row, left to right:}
\textbf{(a)} Photograph of a zebra ($320\times240$ pixels), converted to
greyscale. Original image from
\texttt{http://en.wikipedia.org/wiki/File:Grants\_Zebra.jpg}, author Ruby 1x2,
released to public domain. --
\textbf{(b)} Texture descriptor $I_{f^V}(G^{\mathrm{E}}_{w})$,
$\varrho=7$, $\beta=0.1$, $q=0.1$. --
\textbf{(c)} Texture descriptor $I_{f^V}(T^{\mathrm{E}}_{w})$,
parameters  as in (b). --
\textbf{Middle row, left to right:}
\textbf{(d)} Texture descriptor $I_{f^P}(T^{\mathrm{E}}_{w})$,
parameters as in (b). --
\textbf{(e)} Texture descriptor
$\bar{I}_{\mathrm{D}}^{\mathrm{E}}(T^{\mathrm{A}}_{u})$,
parameters as in (b). --
\textbf{(f)} Initial contour for geodesic active contour segmentation
($t=0$). --
\textbf{Bottom row, left to right:}
\textbf{(g)} Geodesic active contours with edge-stopping function computed
from the texture descriptors $I_{f^P}(T^{\mathrm{E}}_{w})$ shown in (d)
and $\bar{I}_{\mathrm{D}}^{\mathrm{E}}(T^{\mathrm{A}}_{u})$
shown in (e) with $\sigma=7$, Perona-Malik
edge-stopping function, $\lambda=0.014$, $t=100$. --
\textbf{(h)} Same as (g) but $t=400$. --
\textbf{(i)} Same as (g) but $t=1\,340$ (steady state).
}
\label{fi-zebra}
\end{figure}
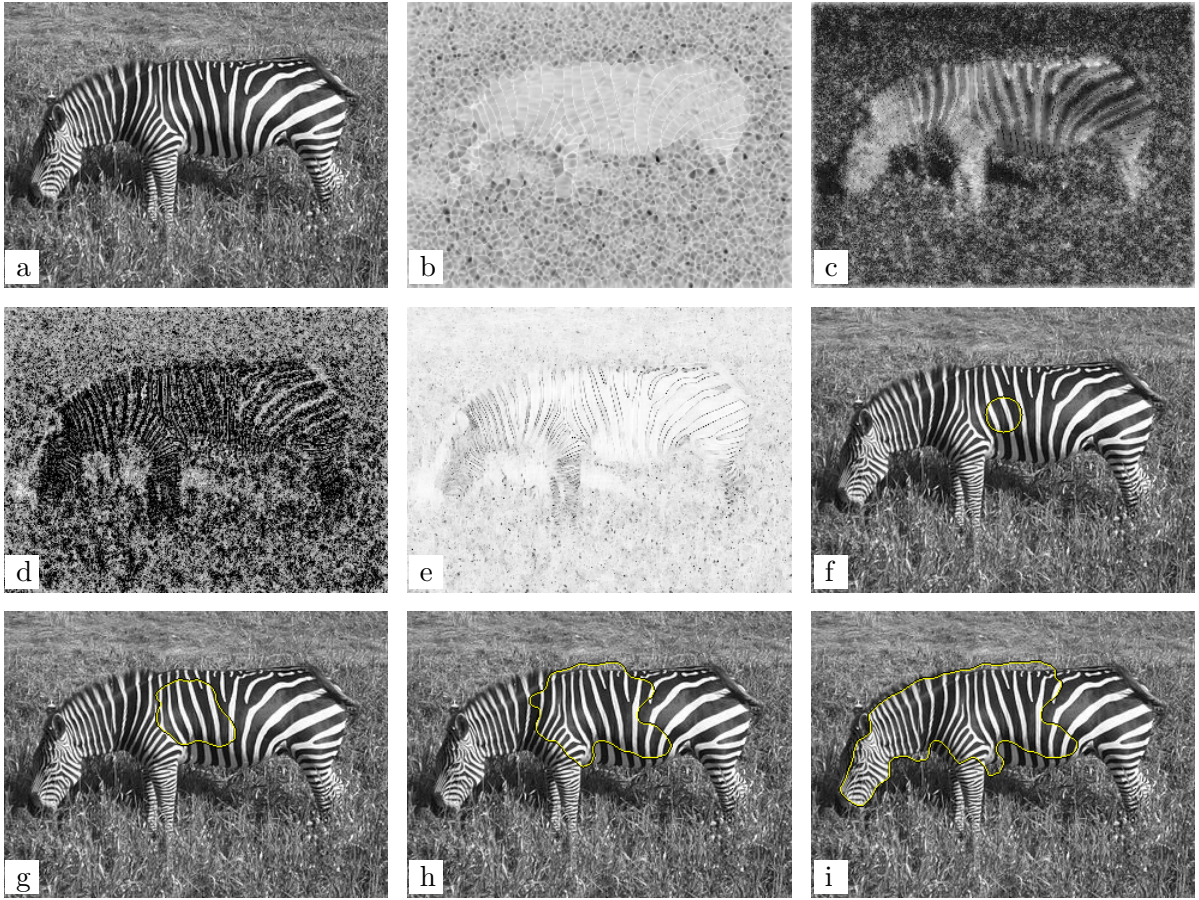

In our last experiment, Figure~\ref{fi-zebra}, we consider a real-world
image showing a zebra, see frame (a). In a sense, this experiment resembles
the synthetic case from Section~\ref{ssec-segex-2} because again a
foreground dominated by a clear stripe pattern is to be distinguished from
a background filled with small-scale detail.
In frames (b)--(e) four texture descriptors are shown. With regard to the
higher resolution of the test image, the patch radius has been chosen
slightly larger than in the previous examples, $\varrho=7$, whereas $\beta=0.1$
was retained.
As can be seen in
frame (b), $I_{f^V}(G^{\mathrm{E}}_{w})$ shows the same kind of
over-segmentation behaviour as observed in Section~\ref{ssec-segex-1},
however, it also separates a large part of the zebra shape well from the
background. The second descriptor, $I_{f^V}(T^{\mathrm{E}}_{w})$ in frame (c),
appears unsuitable here because it does not yield sufficiently similar values
within the black and white stripes to recognise these as a common texture.
In contrast, $I_{f^P}(T^{\mathrm{E}}_{w})$ and
$\bar{I}_{\mathrm{D}}^{\mathrm{E}}(T^{\mathrm{A}}_{u})$
in Figure~\ref{fi-zebra}(d) and (e), respectively, achieve this largely.

Our GAC segmentation in frames (f)--(i) uses a larger Gaussian kernel for
pre-smoothing than before, $\sigma=7$, to flatten out small-scale
inhomogeneities in the texture descriptors, and combines the two descriptors
from (d) and (e). With these data, a large part of the zebra including the
head and front part of the torso is segmented in the final steady state.
Not included are the rear part and the forelegs. Note that in the foreleg
part the stripes are much thinner than in the segmented region, apparently
preventing the recognition of this texture as a continuation of the one from
the head and front torso. In contrast, the rear part of the torso shown
very thick stripes which under the patch size chosen decompose into separate
(homogeneous) textures for black and white stripes, as is also visible in
the texture descriptors (d) and (e) themselves. Further investigation of
parameter variations and inclusion of more texture descriptors might improve
this preliminary result in the future.

\section{Analysis of Graph-Entropy-Based Texture Descriptors}
\label{sec-tdana}

In this section, we undertake an attempt to analyse the texture descriptors
based on the entropy measures $I_{f^V}$ and $I_{f^P}$, focussing on the
question what properties of textures are actually encoded in their
information functionals $f^V$ and $f^P$. Part of this analysis is
on a heuristic level at the present stage of research, and future work
will have to be invested to add precision to these arguments. This applies
to the limiting procedure in Section~\ref{ssec-graphfrac} as well
as to the concept of local fractal dimension arising in
Section~\ref{ssec-fracdim}. We believe, however, that even in its
present shape, the analysis provided in the following gives valuable
intuition about the principles underlying our texture descriptors.

\subsection{Rewriting the Information Functionals}

For the purpose of our analysis, we generalise the information
functional $f^V$ from \eqref{fV-Dehmer-e} directly to
edge-weighted graphs by replacing the series $s_d$ of cardinalities
from \eqref{sd} with the monotone increasing
function $s:[0,\infty)\to\mathbb{R}$,
\begin{equation}
s(d) := \mathrm{vol}(S_d(v_j))
\label{sd-vol}
\end{equation}
that measures volumes of spheres with arbitrary radius. Assuming
the exponential weighting scheme \eqref{expws} and
large $D(G)$ this yields
\begin{align}
f^V(v_i)&\approx\exp\left(\int_0^\infty q^ds(d)\,\mathrm{d}d\right) \;.
\label{fV-edgeweighted-int}
\end{align}
An analogous generalisation of \eqref{fP-Dehmer-e} is
\begin{align}
f^P(v_i)&\approx\exp\left(\int_0^\infty q^dds(d)\,\mathrm{d}d\right) \;.
\label{fP-edgeweighted-int}
\end{align}
%

\subsection{Infinite Resolution Limits of Graphs}
\label{ssec-graphfrac}

We assume now that the image is sampled successively on finer grids,
with decreasing $h_x=h_y=:h$. Note that the number of vertices of any region
of the edge-weighted pixel graph, or any of the derived edge-weighted
graphs introduced in Section~\ref{ssec-graphcons},
grows in this process with $h^{-2}$.
By using the volumes of spheres instead of the original cardinalities,
\eqref{sd-vol} provides a re-normalisation that compensates
this effect in \eqref{fV-edgeweighted-int}.

Thus, it is possible
to consider the limit case $h\to0$. In this limit case, graphs turn into
metric spaces representing the structure of a space-continuous image.
Additionally, these metric spaces are endowed with a volume measure which
is the limit case of the discrete measures on graphs given by vertex counting.

In simple cases, these metric spaces with volume measure can be manifolds.
For example, for a homogeneous grey image without any contrast, the limit of
the edge-weighted pixel graph is an approximation to a plane, i.e.\ a
$2$-dimensional manifold.
For an image with extreme contrasts in one direction, e.g.\ a stripe pattern,
the edge-weighted pixel graphs will be path graphs, resulting in a metric
space as limit which is essentially a $1$-dimensional manifold.
Finally, in the extreme case of a noise image in which neighbouring
pixels have almost nowhere similar grey-values, the graph will practically
decompose into numerous isolated connected components, corresponding
to a discrete space of dimension $0$.

For more general textured images, the limit spaces will possess a more
complicated topological structure. At the same time, it remains possible,
of course, to measure volumes of spheres of different radii in these spaces.
Clearly, sphere volumes will increase with sphere radius. If they fulfil
a power law, the (possible non-integer) exponent can immediately be
interpreted as a dimension. The space itself is then interpreted as
some type of fractal \cite{Mandelbrot-Book77}. The dimension concept
underlying here is almost that of the Minkowski dimension (closely related
to Hausdorff dimension) that is frequently used in fractal theory, with
the difference that the volume measure here is inside the object being
measured instead of in an embedding space.
Based on the above reasoning, values of the dimension will range between
$0$ and $2$.

Note that even in situations in which there is no global power law for the
sphere volumes, and therefore no global dimension, power laws, possibly
with varying exponents, will still be approximated for a given sphere centre
in suitable ranges of the radius, thus allowing to define the fractal
dimension as a quantity varying in space and resolution. This resembles the
situation with most fractal concepts being applied to real-world data: the
power laws that are required to hold for an ideal fractal across all scales
will be found only for certain ranges of scales in reality.

Dijkstra trees, too, turn into $1$-dimensional manifolds
in the case of sharp stripe images; for other cases they will also
yield fractals. Fractal dimensions, wherever applicable, will be below
those observed with the corresponding full edge-weighted pixel graphs, thus,
the relevant range of dimensions is again bounded by $0$ from below
and $2$ from above.

One word of care must be said at this point. The fractal structures obtained
here as limit cases of graphs for $h\to0$ are not identical with the
image manifolds whose fractal structures are studied as a means of texture
analysis in
\cite{Avadhanam-msc93,Pentland-PAMI84,Soille-JVCIR96}
and others. In fact, fractal dimensions of the latter, measured as Minkowski
dimensions by means of the embedding of the image manifold of a grey-scale
image in three-dimensional Euclidean space, range from $2$ to $3$ with
increasing roughness of the image, whereas the dimensions measured in the
present work go down from $2$ to $0$ for increasing image roughness.
Whereas it can be conjectured that these two fractal structures are related,
future work will be needed to gain clarity about this relationship.

\subsection{Fractal Analysis}
\label{ssec-fracdim}

Based on the findings from the previous section, let us now
assume that the limit $h\to0$ from one of the graph structures results
in a measured metric space $F$ of dimension $\delta\in[0,2]$, in which
sphere volumes are given by the equation
\begin{align}
s(d) &= d^\delta U(\delta)
\label{sd-delta}
\end{align}
where
\begin{equation}
U(\delta) =
\frac{\pi^{\delta/2}}{\varGamma(\delta/2+1)}
\label{Udelta}
\end{equation}
is the volume of a unit sphere, $\varGamma$ denoting the Gamma function.
Thus, we assume that
$s(d)$ interpolates the sphere volumes of Euclidean spaces for
integer $\delta$.

Note that this assumption has indeed two parts. The first,
\eqref{sd-delta}, means that a volume measure on the metric
space $F$ exists that behaves homogeneously with degree $\delta$ with
regard to distances. In the manifold case (integer $\delta$),
this is the case of vanishing curvature; for general manifolds of integer
dimension $\delta$, \eqref{sd-delta} would hold as an approximation
for small radii.

The second assumption, \eqref{Udelta}, corresponds to the
Euclideanness of the metric. For edge-weighted pixel graphs based on
$4$- or $8$-neighbourhoods, the volume of unit spheres actually deviates
from \eqref{Udelta}, even in the limit. However, with increasing
neighbourhood size, \eqref{Udelta} is approximated better and better.
Most of the following analysis does not depend specifically on
\eqref{Udelta}; thus we will return to \eqref{Udelta}
only later for numerical evaluation of information functionals.

With \eqref{sd-delta} we have
\begin{align}
\int_0^\infty q^ds(d)\,\mathrm{d}d
&= 
U(\delta)
\int_0^\infty\exp(d\ln q) d^\delta\,\mathrm{d}d
\notag\\*&
= 
(-\ln q)^{\delta+1} 
U(\delta)
\int_0^\infty\exp(-w)w^\delta\,\mathrm{d}w
\notag\\*&
= (-\ln q)^{\delta+1} 
U(\delta)
\varGamma (\delta+1)
\end{align}
where we have substituted $w:=-d\ln q$.
As a result, we obtain
\begin{align}
f^V(v_i)&\approx\exp\left((-\ln q)^{\delta+1} U(\delta) \varGamma(\delta+1)
\right)
\;.
\label{fV-frac}
\end{align}
Analogous considerations for $f^P$ from \eqref{fP-edgeweighted-int}
lead to
\begin{align}
f^P(v_i)&\approx\exp\left((-\ln q)^{\delta+2} U(\delta) \varGamma(\delta+2)
\right)
\;.
\label{fP-frac}
\end{align}
As pointed out before,
the metric structure of the fractal $F$ will in general
be more complicated such that it does not possess a well-defined
global dimension. However, such a dimension can be measured at each
location and scale.
The quantities $f^V$ and $f^P$ as stated in \eqref{fV-frac},
\eqref{fP-frac} can then be understood as functions of the local
fractal dimension in a neighbourhood of vertex $v_i$ where the size of
the neighbourhood -- the scale -- is controlled by the decay of the
function $q^d$ in the integrands of \eqref{fV-edgeweighted-int}
and \eqref{fP-edgeweighted-int}, respectively.

As a result, we find that the information functionals $f^V$ and $f^P$
represent distribution over the input pixels of an image patch (non-adaptive
or adaptive) in which the pixels are assigned different weights dependent
on a local fractal dimension measure. The entropies $I_{f^V}$ and $I_{f^P}$
then measure the local homogeneity or inhomogeneity of this dimension
distribution: For very homogeneous dimension values within a patch, the
density resulting from each of the information functionals $f^V$,
$f^P$ will be fairly homogeneous, implying high entropy. The more
the dimension values are spread out, the more will the density be dominated
by a few pixels with high values of $f^V$ or $f^P$, thus yielding low
entropy. The precise dependency of the entropy on the dimension distribution
will be slightly different for $f^V$ and $f^P$ and will also depend on the
choice of $q$. Details of this dependency will be a topic of future work.

\begin{figure}[t!]
\unitlength0.001\textwidth
\begin{picture}(1000, 505)
\put(  5,500){\rotatebox{270}{\includegraphics[height=350\unitlength]{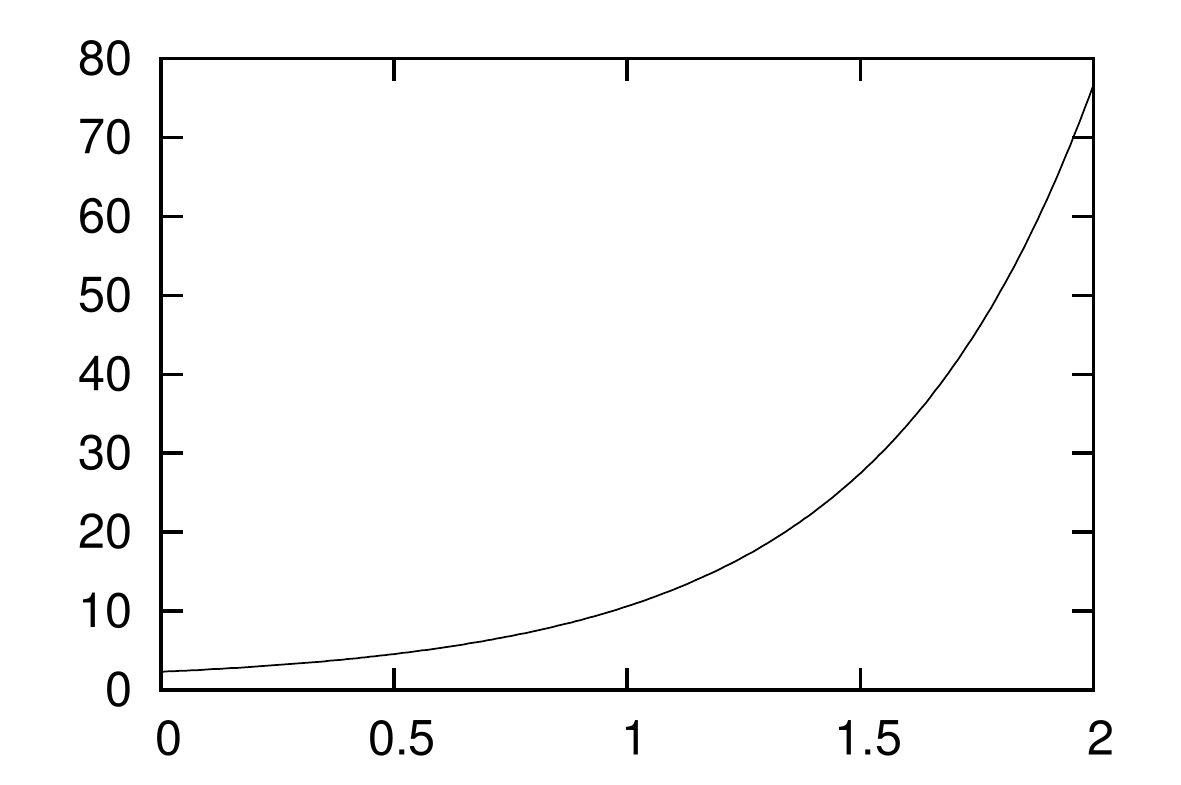}}}
\put(  0,370){\rotatebox{90}{\tiny$\ln f^V$}}
\put(190,260){\tiny$\delta$}
\put( 15,260){(a)}
\put(333,500){\rotatebox{270}{\includegraphics[height=350\unitlength]{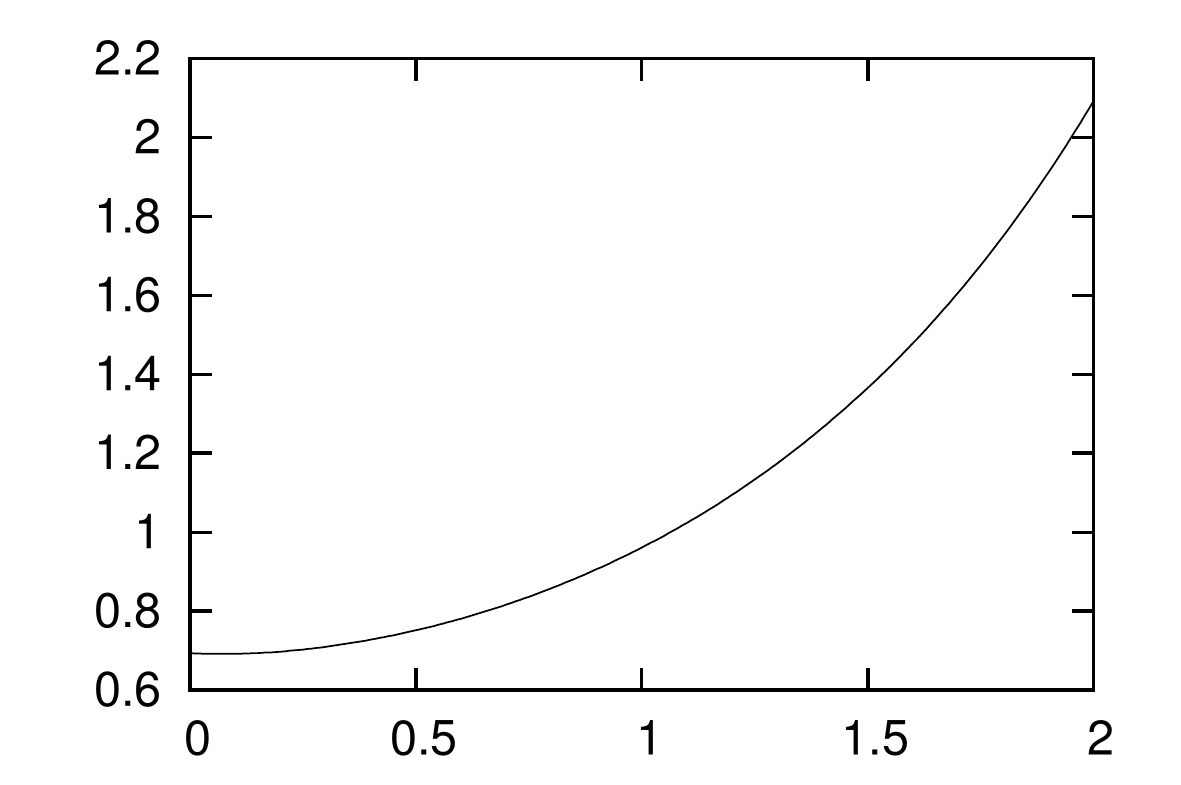}}}
\put(338,370){\rotatebox{90}{\tiny$\ln f^V$}}
\put(525,260){\tiny$\delta$}
\put(350,260){(b)}
\put(661,500){\rotatebox{270}{\includegraphics[height=350\unitlength]{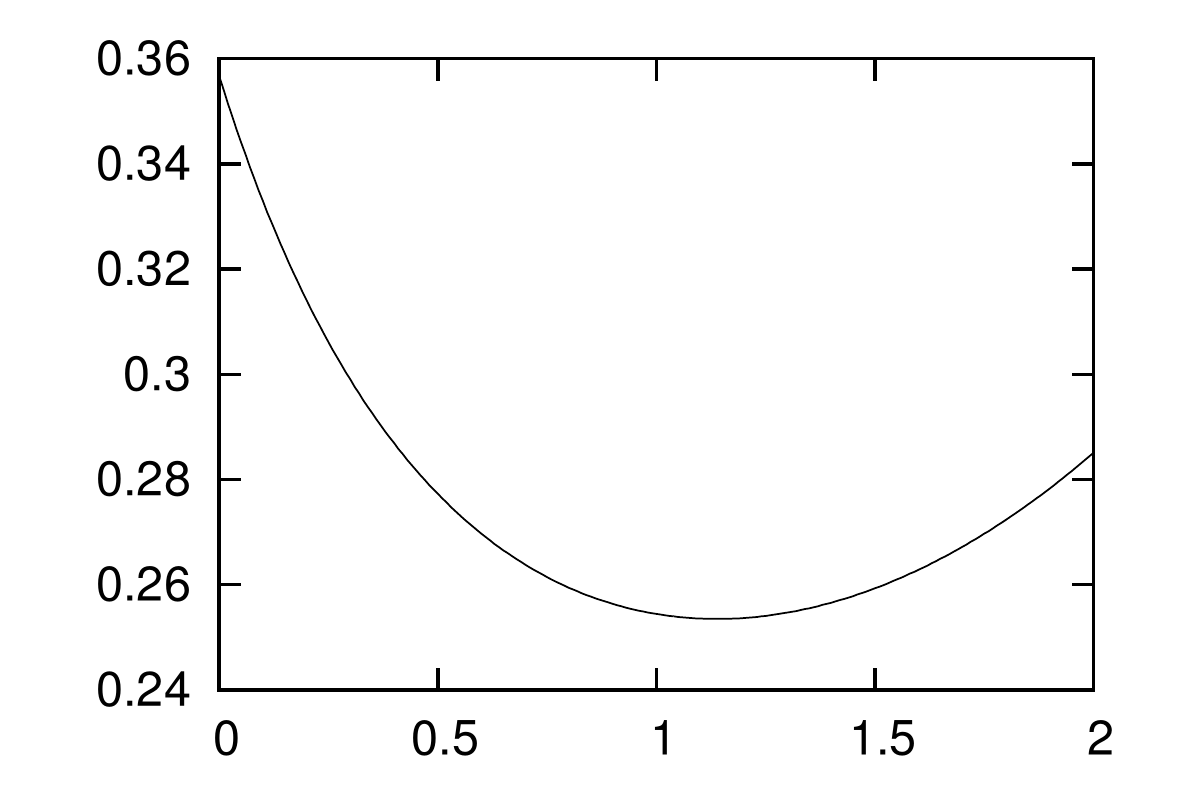}}}
\put(666,370){\rotatebox{90}{\tiny$\ln f^V$}}
\put(853,260){\tiny$\delta$}
\put(680,260){(c)}
\put(  5,240){\rotatebox{270}{\includegraphics[height=350\unitlength]{ifv-01.pdf}}}
\put(  0,110){\rotatebox{90}{\tiny$\ln f^P$}}
\put(190,  0){\tiny$\delta$}
\put( 15,  0){(d)}
\put(333,240){\rotatebox{270}{\includegraphics[height=350\unitlength]{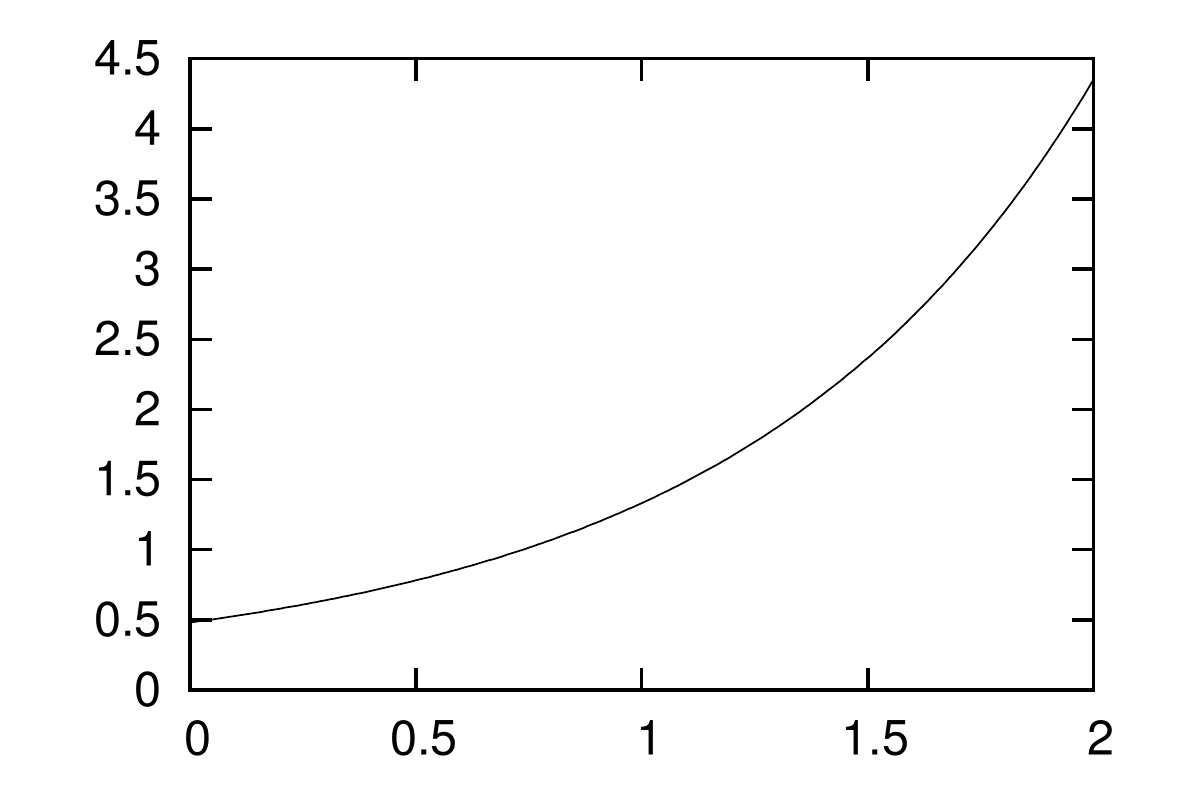}}}
\put(338,110){\rotatebox{90}{\tiny$\ln f^P$}}
\put(525,  0){\tiny$\delta$}
\put(350,  0){(e)}
\put(661,240){\rotatebox{270}{\includegraphics[height=350\unitlength]{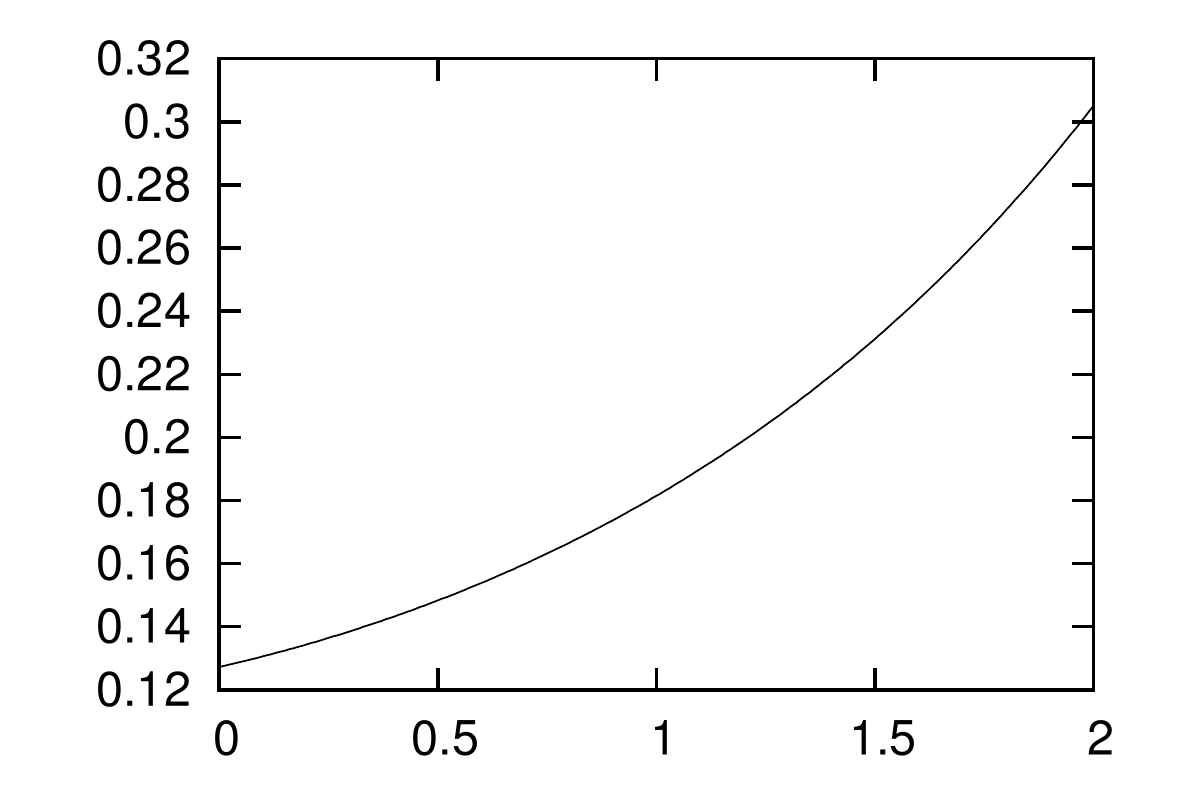}}}
\put(666,110){\rotatebox{90}{\tiny$\ln f^P$}}
\put(853,  0){\tiny$\delta$}
\put(680,  0){(f)}
\end{picture}
\caption{%
Information functionals (in logarithmic scale) as functions of local dimension.
\textbf{Top row:} $\ln f^V$. \textbf{Left to right:}
\textbf{(a)} $q=0.1$. --
\textbf{(b)} $q=0.5$. --
\textbf{(c)} $q=0.7$. --
\textbf{Bottom row:} $\ln f^P$. \textbf{Left to right:}
\textbf{(d)} $q=0.1$. --
\textbf{(e)} $q=0.5$. --
\textbf{(f)} $q=0.7$.
}
\label{fi-fVfP}
\end{figure}

To give a basic intuition, we present in Figure~\ref{fi-fVfP}
graphs of $\ln f^V$ and $\ln f^P$ as functions of the dimension
$\delta\in[0,2]$ for selected values of $q$. In computing these values,
the specific choice \eqref{Udelta} for $U(\delta)$ has been used.

In the left column, frames (a) and (b), we use $q=0.1$ as in our
experiments in Section~\ref{sec-segex}. Here, both $\ln f^V$ and
$\ln f^P$ increase drastically by almost $10$ from $\delta=0$ to $\delta=1$
and even by $70$ from $\delta=1$ to $\delta=2$. In the resulting
information functionals $f^V$ and $f^P$, i.e.\ after applying $\exp$ to
the functions shown in the figure, even pixels with only slightly higher
values of the dimension strongly dominate the entire information density
within the patch.

For increasing $q$, the rate of increment in $f^V$ and $f^P$ with $\delta$
becomes lower. For $q=0.5$ as shown in the second column, frames (b) and
(e), of Figure~\ref{fi-fVfP}, the variation of $\ln f^V$ and
$\ln f^P$ is already reduced to $2$ and $4$, respectively, such that vertices
across the entire dimension range $[0,2]$ will have a relevant influence
on the information density. For even larger $q$ the dependency of
$f^V$ and $f^P$ on $\delta$ becomes non-monotonic (as shown in (c)
for $f^V$ with $q=0.7$) and even monotonically decreasing (for both
$f^V$ and $f^P$ at $q=0.9$; not shown).
It will therefore be interesting for further investigation to evaluate
also the texture
discrimination behaviour of the entropy measures for varying $q$,
as this may be a way to targeting the sensitivity of the measures
specifically at certain dimension ranges.

In this context, however, it becomes evident that the parameter $q$
plays two different roles at the same time. First, it steers the
approximate radius of influence for the fractal dimension estimation. Here, it
is important that this radius of influence is smaller than the patch size
underlying the graph construction, such that the cut-off of the graphs has
no significant influence on the values of the information functional at the
individual vertices. Second, $q$ determines the shape and steepness of the
function (compare Figure~\ref{fi-fVfP}) that relates the local
fractal dimension to the values of the information functionals.
This makes it desirable to refine in future work the parametrisation of the
exponential weighting scheme \eqref{expws} so that the two roles
of $q$ are distributed to two parameters.

\section{Conclusion}
\label{sec-conc}

In this paper, we have presented the framework of graph-index-based
texture descriptors that has first been introduced in
\cite{Welk-qgt14}. Particular emphasis was put on entropy-based
graph indices that have proven in \cite{Welk-qgt14} to afford
medium to high sensitivity for texture differences.

We have extended the work from \cite{Welk-qgt14} in two directions.
Firstly,
we have stated an approach to texture-based image segmentation in which
the texture descriptor framework was integrated with geodesic active
contours \cite{Caselles-iccv95,Kichenassamy-iccv95}, a
standard method for intensity-based image segmentation. This approach
was already briefly introduced in \cite{Welk-bookchapter15x}
and is demonstrated here by a larger set of experiments, including
two synthetic and one real-world example. Secondly, we have analysed
one representative of the graph-entropy-based texture descriptors
in order to gain insight about the image properties that this descriptor
relies on. It turned out that it stands in close relation to measurements
of fractal dimension of certain metric spaces that arise from the
graphs in local image patches that underly our texture descriptors.
Although this type of fractal dimension measurement in images differs
from existing applications of fractal theory in image (and texture) analysis,
as the latter treat the image manifold as fractal object, results indicate
that the two fractal approaches are related.

Our texture descriptor framework as a whole and also both
novel contributions presented here require further research.
To mention some topics, we start with parameter selection of
the texture descriptors. In \cite{Welk-qgt14,Welk-bookchapter15x}
as well as in this paper, most parameters were fixed to specific values
based on heuristics. A systematic investigation of the effect of variations
of all these parameters is part of ongoing work. Inclusion of novel
graph descriptors proposed in the literature, e.g.\
\cite{Dehmer-AMC15}, is a further option.

The algorithms currently used for the computation of graph-entropy-based
texture descriptors
need computation times in the range of minutes already for small images.
As the algorithms have not been designed for efficiency, there is much room
for improvement which will also be considered in future work.

Both texture discrimination and texture segmentation have been demonstrated
so far rather on a proof-of-concept level. Extensive evaluation on larger
sets of image data is ongoing. This is also necessary to gain more insight
about the suitability of particular texture descriptors from our set for
specific classes of textures.

Regarding the texture segmentation framework, the conceptual break between
the graph-based set of texture descriptors and the partial differential
equation for segmentation could be reduced by using e.g.\ a graph-cut
segmentation method. It can be asked whether such a combination even allows
for some synergy between the computation steps. This is not clear so far
since the features used to weight graph edges are different: intensity
contrasts in the texture descriptor phase; texture descriptor differences
in the graph-cut phase. In further course, the integration of
graph-entropy-based texture descriptors into more complex segmentation
frameworks will be a goal. Unsupervised segmentation approaches are
not capable to handle involved segmentation tasks (like in medical diagnostics)
where highly accurate segmentation can only be achieved by including
prior information on the shape and appearance of the objects to be segmented.
State-of-the-art segmentation frameworks therefore
combine the mechanisms of unsupervised segmentation approaches with
model-based methods as introduced e.g.\ in \cite{Cootes-tr01}.

On the theoretical side, the analysis of the fractal limit of the descriptor
construction will have to be refined and extended to include all six
graph settings from Section~\ref{ssec-graphcons}. Relations between
the fractal structures arising from the graph construction and the more
image manifold more commonly treated in fractal-based image analysis will
have to be analysed. Generally, much more theoretical work deserves to be
invested in understanding the connections and possible equivalences
between the very disparate approaches to texture description that can be
found in the literature.
A graph-based approach like ours admits different directions of such
comparisons. It can thus be speculated that it could play a pivotal role in
understanding the
relations between texture description methods, and create a unifying view
on different methods that would also have implications for the understanding
of texture itself.


\end{document}